%% file: main.tex
\documentclass[manuscript]{acmart}
\newcommand{\cd}[1]{\texttt{#1}}
\usepackage{url,hyperref,microtype,subcaption}
\usepackage{siunitx}
\usepackage{subcaption}
\usepackage{pgfplotstable}
\usepackage{booktabs}
\usepackage{xr-hyper}
\usetikzlibrary{arrows.meta}

\title{Small-scale neural simulation of
\textit{Caenorhabditis elegans}
on Loihi-2 using NeuroML2Loihi}

\author{Danny Rosen}
\email{daniel20@vt.edu}

\affiliation{
  \department{Institute for Advanced Computing}
  \institution{Virginia Polytechnic Institute and State University}
  \city{Alexandria}
  \state{Virginia}
  \country{USA}
}

\author{Ramashish Gaurav}
\affiliation{
  \department{Institute for Advanced Computing}
  \institution{Virginia Polytechnic Institute and State University}
  \city{Alexandria}
  \state{Virginia}
  \country{USA}
}

\author{Yang Yi}
\affiliation{
  \department{Institute for Advanced Computing}
  \institution{Virginia Polytechnic Institute and State University}
  \city{Alexandria}
  \state{Virginia}
  \country{USA}
}

\begin{document}

\input{abstract}

\maketitle
\input{introduction}

\input{background_and_related_work}

\section{Methods \label{sec:methods}}

\subsection{NeuroML2Loihi}
\label{sec:n2l}
NeuroML2Loihi (N2L) is a novel python framework for the conversion of NeuroML-formatted specifications of cognitive models into Lava Loihi primitives.  This tool creates an interface between the State-of-the-Art in neuromorphic hardware and standardized descriptions of cognitive models.  Such a connection facilitates direct comparison between the Loihi~2 platform, traditional HPC simulations tools, as well as existing neuromorphic platforms such as BrainScaleS2 and Spinnaker which are already compatible with \cd{NeuroML} through \cd{PyNN}.  NeuroML2Loihi is released as open-source software and is available at \url{https://github.com/DRosen766/NeuroML2Loihi}. The archived version used in this work is available on Zenodo (DOI: 10.5281/zenodo.21431945)~\cite{dannyrosenDRosen766NeuroML2LoihiNeuroML2Loihi2026}.  The following sections describe the key features of the framework.

\input{diagrams/fig_architecture}

\subsubsection{Configuration File Parsing}
Each N2L experiment is described by \textit{two} xml-formatted files in tandem.  The first, an \texttt{nml} file, describes elements of the neural model including neurons, synapses, external stimuli, and the parameters that define each.  These files are parsed by N2L's \texttt{NMLParser} and produce a \texttt{Network} object.  Secondly \textit{LEMS} files describe the simulation environment including duration, time-step length, and neurons to record.   These files are parsed by N2L's \texttt{LEMSParser} to produce a \texttt{Config} object.

\subsubsection{Mappings}
 NeuroML2Loihi uses a mapping layer to associate NeuroML tags with their corresponding Lava process implementations. When \texttt{NML} files are parsed in the \texttt{Network} module, each supported NeuroML element is resolved through this mapping dictionary before the corresponding Lava process is initialized. For example, an \texttt{iafCell} tag can be mapped to a Loihi-compatible \texttt{LIF} process, while an \texttt{expTwoSynapse} tag can be mapped to a \texttt{Dense} synaptic process. This separates NeuroML parsing from Lava-specific implementation choices and makes the framework extensible: users can add support for additional NeuroML elements or custom Lava processes by registering new tag-process pairs through \cd{register\_mapping()}.

\subsubsection{Networks}
N2L's \texttt{Network} class leverages the NetworkX~\cite{hagbergExploringNetworkStructure2008} library to form an intermediate representation of the neural model that has been parsed from \texttt{nml} and \texttt{LEMS} files.  The parsed \texttt{nml} file is converted into labeled neurons and weighted connections.  From this intermediate representation, Lava neuron \texttt{Process}es (ex. \texttt{LIF}) are created according to the number of neurons in the model.  Synaptic \texttt{Process}es (ex. \texttt{Dense}) are then initialized to represent neuron connections using an adjacency matrix derived from the edge weights of the \texttt{networkx} graph.  The output of the neuron \texttt{Process} is fed into the synaptic \texttt{Process}.  In turn, the output of the synaptic \texttt{Process} is fed as input into the neuron \cd{Process}, thus creating a recurrent network.  Additional \cd{Process}es are created for input and recording purposes as described below.
\input{diagrams/fig1_lava_network}
\subsubsection{Experiments}
In the context of \texttt{N2L}, an experiment refers to an atomic simulation of an arbitrary duration on a single set of configuration files.  The \texttt{Experiment} class captures functionality for input generation, spike/voltage recording, and execution.  Based on the input pattern specified in the \texttt{LEMS} file, \texttt{SpikeBuffers} are created to send spikes via \texttt{Dense} \cd{Process} to the neuron \cd{Process}.  For recording on simulation, \cd{Monitor} \cd{Process}es are used to track voltage and spike data.  For hardware recording, a \cd{Dense} \texttt{Process} is connected to the output of the neuron \cd{Process} with a connection matrix to route spiking outputs of recorded neurons to \cd{Receiver} \cd{Process}es.  \cd{StateProbes} are then used to track spiking outputs for each \cd{Receiver} at each timestep.  For voltage recording on Loihi-2 hardware, a \cd{StateProbe} is connected to the neuron \cd{Process}'s \cd{v} property.  Once all components are initialized and connected, simulations are executed for the specified number of timesteps.

\input{diagrams/fig2_hardware_pipeline}

\subsubsection{Comparison}
Results of experiments are saved as json and .dat files.  The .dat files share the exact format of the .dat files produced by c302 simulation.  The N2L framework natively supports the comparison of spike trains using \textit{Van Rossum} \cite{van2001novel} and \textit{Victor-Purpura} \cite{victor1996nature} distance metrics.  Additionally, N2L supports \textit{Cross Correlation entropy} on voltage traces.  The algorithm implementations are based on \cite{elephant18}.

\subsubsection{Extensibility}
\cd{N2L} is designed to maximize extensibility and customizability to enable granular experimentation.  To accomplish this, users are able to specify which Lava \cd{Process}es correspond to which components in the \cd{nml} configuration through a dynamic mapping registration component.  This permits the use of the same configuration for experimentation with various neuron and synapse models.  Additionally, users may implement their own Lava-native neuron models to include in their experiments. Crucially, access to Loihi~2 hardware is not necessary to utilize the \cd{N2L} framework.  Users can easily switch between Loihi~2 hardware and simulator backends. Note that Loihi's simulator backend runs entirely on CPU and does not require a physical Loihi~2 device.

\input{experiment_configurations}
\input{benchmarking_methods}
\input{openworm_simulation_comparison}

\section{Results}
\label{sec:results}
\input{similarity_results}
\input{benchmarking_metrics_results}
\section{Discussion}
\label{sec:discussion}
\input{benchmarking_analysis_discussion}
\input{conclusion}

\bibliographystyle{ACM-Reference-Format}
\bibliography{test}
\end{document}

%% file: abstract.tex
\begin{abstract}
Neural simulation is an essential methodology for studying biological nervous systems, but it remains limited by the computational costs of repeatedly updating neural and synaptic states over many timesteps. Neuromorphic hardware is a promising substrate for this class of workloads because such devices combine event-driven execution, localized memory, and spike-based communication.  However, neuromorphic \textit{software} remains comparatively fragmented and difficult to integrate with the model-description standards used by the computational neuroscience community. This disconnect limits reproducible experimentation and makes direct comparison between conventional and neuromorphic simulation workflows difficult. This paper introduces ``NeuroML2Loihi'', an open-source framework for translating NeuroML model descriptions into Loihi~2-deployable Lava primitives, enabling standardized biological models to be executed on Intel's Loihi~2 neuromorphic chip. We evaluate our framework using c302-derived \textit{C. elegans} neural models across three connectome-scale configurations. Cross-platform similarity is assessed using spike-train and voltage-trace metrics, while latency, power, and energy consumption are benchmarked against the conventional reference workflow. Across the evaluated models, Loihi~2 execution enabled through NeuroML2Loihi achieves 514--1211$\times$ lower latency and a 1.8--1.9$\times$ reduction in mean wattage relative to the NEURON reference, producing substantially lower total energy consumption primarily through reduced execution time. These results demonstrate that standards-based neuromorphic neural simulation is feasible on Loihi~2 and can support meaningful comparison against established computational neuroscience workflows.
\end{abstract}

%% file: introduction.tex
\section{Introduction}
Neural simulation plays a key role in the study of biological nervous systems because it allows experimentation on neural models without invasive procedures on living organisms. As model descriptions become more detailed, simulation provides a practical path for connecting anatomical and physiological data to hypotheses about neural activity and circuit behavior. However, neural simulation remains a difficult computational workload. Biological neural networks require repeated updates of neuron state, synaptic state, and dense inter-neuron communication over many timesteps. These requirements create pressure on latency, memory occupancy, and energy consumption. Achieving real-time execution is difficult, storing and updating neural and synaptic states can become expensive, and repeated simulation over many seeds or parameter settings can make runtime and energy costs practically limiting. Thus, although the conventional High Performance Cluster (HPC) infrastructure has a more mature software ecosystem, broader tooling support, and a longer history of deployment for computational neuroscience workloads, they struggle in terms of minimizing energy costs and runtime latency; and therefore, struggle to scale.

Neuromorphic hardware is a promising candidate for addressing these challenges. Neuromorphic hardware architectures implement principles from biological nervous systems, including event-driven execution, spike-based communication, and localized memory. These features directly target several of the bottlenecks that make neural simulation expensive: repeated state updates, communication among sparse recurrent networks, and the movement of neural and synaptic state through conventional memory hierarchies. Therefore, neuromorphic devices may be advantageous for neural simulation workloads.
Despite this potential, neuromorphic hardware has not become a standard platform for computational neuroscience simulation. One major reason is the relative fragmentation of the neuromorphic software ecosystem. Traditional neural simulation is supported by mature simulators and acceleration stacks. By contrast, neuromorphic platforms' infrastructure for neural simulation remains comparatively less mature and often rely on device-specific programming models and narrower deployment workflows. This limits adoption, standardized comparison, and progress toward reusable neuromorphic simulation infrastructure.

This gap is especially important for Intel's Loihi~2 neuromorphic chip. Loihi~2 is a digital neuromorphic research processor \cite{orchard2021efficient}, and Lava is the native software framework for constructing and deploying applications to Loihi~2-compatible neuromorphic targets. However, Lava exposes hardware-oriented abstractions, which can be difficult to use from the perspective of computational neuroscience workflows, where models are often specified through simulator-level or standard-description formats. Existing software platforms address important parts of neuromorphic application development, but they do not provide a direct route from established neuroscience model-description standards to Loihi~2 execution.


To address this gap, this paper introduces ``NeuroML2Loihi'', a software framework for executing NeuroML-formatted neural models on Loihi~2 hardware. NeuroML2Loihi (herein referred to as N2L) parses NeuroML and LEMS files, constructs an intermediate network representation, maps the model to Loihi~2-compatible Lava primitives, executes the simulation on Loihi-2 hardware or simulator backends, and records outputs for comparison against reference simulations. NeuroML is selected as the source standard because it is an established, XML-based model-description language for computational neuroscience~\cite{gleeson2010neuroml}. It is also used by the OpenWorm ecosystem for modeling the \textit{Caenorhabditis elegans} (\textit{C. elegans}) nematode nervous system~\cite{szigeti2014openworm,gleeson2018c302}.  Therefore, NeuroML2Loihi allows neuromorphic hardware to serve as a backend for an already-defined OpenWorm simulation.  N2L's output artifact format also mirrors that of the OpenWorm simulation to allow for easy and direct comparison between backends.

As a case study, we deploy c302-derived models of the \textit{C. elegans} nervous system onto a single Loihi~2 chip. \textit{C. elegans} is a useful target because its nervous system is small enough to be tractable while still representing a full neuron set. The full model evaluated in this work contains 302 neurons, allowing the experiment to move beyond isolated toy networks while remaining within the resources of a single Loihi-2 chip~\cite{whiteStructureNervousSystem1986,varshneyStructuralPropertiesCaenorhabditis2011}.  To the best of the authors' knowledge, this is the first deployment of a \textit{C. elegans} neural model including all $302$ neurons on a single neuromorphic chip.

This work evaluates NeuroML2Loihi along two axes: first, we assess whether the Loihi~2-based simulation preserves neural behavior relative to an NEURON reference simulation using spike-train and voltage-trace metrics, second, we benchmark latency, memory occupancy, power, and energy consumption across the Loihi~2 implementation and the conventional reference workflow. The goal is not only to demonstrate that a NeuroML-derived model can run on Loihi~2, but also to determine whether the resulting simulation is accurate enough for meaningful comparison and is efficient enough to motivate neuromorphic deployment.
The contributions of this paper are as follows:
\begin{enumerate}
    \item We introduce ``NeuroML2Loihi'', a software framework for executing NeuroML model descriptions on Loihi-2 neuromorphic chip through Loihi~2-deployable Lava primitives.
    \item We demonstrate deployment of a full 302-neuron \textit{C. elegans} neural model on a single Loihi~2 chip.
    \item We assess cross-platform similarity between Loihi-based simulation and an OpenWorm/NEURON reference simulation using spike-train and voltage-trace metrics.
    \item We benchmark latency, memory occupancy, power, and energy consumption across Loihi 2 and the conventional OpenWorm reference workflow.
\end{enumerate}
The remainder of this paper is organized as follows. Section~\ref{sec:background-related-work} provides background and related work on neural simulation hardware, traditional and neuromorphic simulation software, standardized model descriptions, and prior neuromorphic neural simulation efforts. Section~\ref{sec:methods} 
presents the NeuroML2Loihi framework, and describes the experimental models, the Loihi~2 deployment, benchmark metrics collection and the NEURON reference configuration. Section~\ref{sec:results} reports cross-platform similarity and benchmarking results. Section~\ref{sec:discussion} discusses the interpretation and implications of the results for larger-scale neural models on neuromorphic hardware; followed by Section \ref{sec:conc-fw}, which concludes this study and identifies future work.

%% file: background_and_related_work.tex
\section{Background and Related Work}
\label{sec:background-related-work}

\subsection{Neural Simulation}
\label{sec:neural-simulation}

Significant effort has been devoted to the simulation of neural systems on conventional high-performance computing infrastructure. The Blue Brain Project is a prominent example of this direction, using supercomputing resources to construct biologically detailed cortical models~\cite{markramBlueBrainProject2006}. Other large-scale efforts have explored petascale spiking-network simulation, K-computer-scale network simulation, and cerebellar simulation~\cite{heliasSupercomputersReadyUse2012,kunkelSpikingNetworkSimulation2014, LargestNeuronalNetwork,yamazakiHumanscaleBrainSimulation2021}. These works show that large-scale neural simulation is possible, but that memory, latency, real-time execution, energy consumption, and lack of complete biological ground truth remain central barriers.

Smaller organisms provide a practical path for developing solutions to surpass these limitations. Their nervous systems are more tractable to map, simulate, and compare against experimental observations while still supporting biologically meaningful circuit dynamics. The nematode \textit{Caenorhabditis elegans} is especially important in this context. Brenner emphasized the value of \textit{C. elegans} as a model organism, and its small nervous system has made it a standard substrate for linking neural structure to behavior~\cite{brennerNaturesGiftScience2003}. The adult hermaphrodite nervous system contains only 302 neurons, but it supports sensory and locomotory behaviors, including touch-mediated responses with experimentally characterized circuitry~\cite{whiteStructureNervousSystem1986,chalfieNeuralCircuitTouch1985,riddleMechanosensoryControlLocomotion1997a,varshneyStructuralPropertiesCaenorhabditis2011}.  Additionally, the \textit{C. elegans} nervous system exhibits substantial neuronal diversity despite containing only 302 neurons~\cite{hobertRevisitingNeuronalCell2016}. Recent connectome analyses further demonstrate that \textit{C. elegans} remains an active substrate for studying circuit structure and neuronal function~\cite{emmonsComprehensiveAnalysisElegans2024}.

The OpenWorm project builds on this foundation by providing an open-source ecosystem for modeling \textit{C. elegans}\cite{szigeti2014openworm}. Within this ecosystem, c302 provides a framework for generating neural models of the \textit{C. elegans} nervous system\cite{gleeson2018c302}. OpenWorm and c302 are especially relevant here because they use NeuroML to describe neural models and simulation configurations. NeuroML provides a standardized XML-based language for describing biologically detailed models of neurons and networks~\cite{gleesonNeuroMLLanguageDescribing2010a}, while libNeuroML and PyLEMS support Python-based manipulation and execution of these declarative model descriptions~\cite{vellaLibNeuroMLPyLEMSUsing2014}. Existing OpenWorm/c302 workflows integrate naturally with conventional simulators such as NEURON~\cite{hinesNEURONSimulationEnvironment1997}, and CoreNEURON extends this ecosystem with optimized execution on modern CPU and GPU systems~\cite{kumbharCoreNEURONOptimizedCompute2019}. However, this mature conventional path does not yet provide direct deployment onto Intel's Loihi 2.

\subsection{Neural Simulation on Neuromorphic Hardware}
\label{sec:neuromorphic-simulation}

Neuromorphic hardware is a natural candidate for neural simulation because it is designed around event-driven execution, localized state, spike-based communication, and parallel neural updates~\cite{indiveriImportanceSpaceTime2019,kudithipudiNeuromorphicComputingScale2025a}. Loihi 2 is 
Intel's recent digital neuromorphic research processor (discussed in further detail in Section \ref{subsubsec:loihi-2}) for programmable, event-driven spiking neural computation, and Lava provides the native software framework for constructing and deploying neural models to Loihi~2-compatible neuromorphic targets~\cite{NextLevelNeuromorphicComputing}. These properties make Loihi 2 attractive for neural simulation workloads, particularly when spiking activity is sparse and recurrent communication can be kept close to the compute substrate.

Prior work has demonstrated several forms of neuromorphic neural simulation. Loihi~1 (a previous Loihi device) has been used for simplified \textit{C. elegans}-inspired klinokinesis, showing that worm-inspired neural computation can be adapted to neuromorphic constraints~\cite{kishoreSimplifiedKlinokinesisUsing2021}. BrainScaleS-2 has also been used to emulate insect-inspired circuits for neuromorphic navigation~\cite{schreiberEmulatingInsectBrains2024}. These works demonstrate the feasibility of biologically inspired neuromorphic control, but they focus on reduced or task-specific circuits rather than deployment of a complete neuron set from a biological model.  Neither deploy a full neuron set.

Larger neuromorphic simulations have also been explored. SpiNNaker has been compared with NEST for a full-scale cortical microcircuit model, exposing tradeoffs among timestep accuracy, runtime, power, and energy~\cite{vanalbadaPerformanceComparisonDigital2018}. Later work demonstrated real-time cortical simulation on SpiNNaker and reported energy-to-solution benefits relative to conventional HPC systems~\cite{rhodesRealtimeCorticalSimulation2019}. This is an important point of comparison for neuromorphic neural simulation, but it remains a cortical microcircuit benchmark rather than the deployment of a complete animal nervous system extracted from a biological organism. Most recently, Felix Wang et. al presented a notable large-scale Loihi 2 deployment of the adult \textit{Drosophila melanogaster} connectome, mapping the FlyWire model of roughly 140K neurons and 50M synapses onto 12 Loihi 2 chips~\cite{dorkenwaldNeuronalWiringDiagram2024,wangNeuromorphicSimulationDrosophila2025}. This result provides strong evidence that neuromorphic hardware is becoming capable of supporting biologically realistic connectome-scale simulation, and it complements the present work by illustrating the potential of Loihi~2 at larger biological scales. 

\subsection{Neural Simulation Software}
\label{sec:simulation-software}

These neuromorphic results show clear progress at the hardware level, but turning that progress into reusable neural-simulation workflows also depends on software support. Conventional simulators such as NEURON~\cite{hinesNEURONSimulationEnvironment1997}, NEST~\cite{gewaltigNESTNEuralSimulation2007a}, and Brian/Brian2~\cite{goodmanBrianSimulatorSpiking2008,stimbergBrian2Intuitive2019a} provide established environments for specifying and executing neural models, while acceleration stacks such as CUDA~\cite{CUDAProgrammingGuide} and ROCm~\cite{AMDROCmDocumentation} allow these workflows to target widely available CPU and GPU hardware. NEURON and CoreNEURON are especially relevant here because they provide the conventional reference path used by OpenWorm/c302-style models~\cite{hinesNEURONSimulationEnvironment1997,kumbharCoreNEURONOptimizedCompute2019}. NEST has similarly been used for large-scale spiking-network simulation on HPC systems~\cite{heliasSupercomputersReadyUse2012,kunkelSpikingNetworkSimulation2014}.

Several projects have attempted to improve portability across simulators and model representations. PyNN provides a common interface for neuronal network simulators and has supported workflows across conventional and neuromorphic backends such as SpiNNaker and BrainScaleS~\cite{davisonPyNNCommonInterface2008, furberOverviewSpiNNakerSystem2013, vanalbadaPerformanceComparisonDigital2018}. Neurons-to-Algorithms approaches modeling as a translation problem across levels from neurons to algorithms~\cite{rothgangerN2AComputationalTool2014}. These systems are related to NeuroML2Loihi because they recognize that simulation workflows should \textit{not} be tied to one execution substrate. However, they do not provide a direct open-source  NeuroML-to-Loihi 2 deployment path for complete biological neuron sets.

Thus, we find that neuromorphic software remains less unified. This is partly due to hardware diversity: neuromorphic systems differ in neuron primitives, synapse representations, routing mechanisms, memory organization, timing behavior, and programming interfaces. As a result, neuromorphic workflows often require platform-specific expertise. NengoLoihi and Lava provide important Loihi-family development paths, but they do not directly allow existing NeuroML models to be executed on Loihi 2 while preserving comparison against conventional simulator outputs \cite{NengoLoihiNengoLoihi111dev0}. Brian2Lava is better suited to computational neuroscience workflows, but it lacks a native mechanism for deploying standards-based model descriptions \cite{negriBrian2Lava2025}.  Additionally, Brian2Loihi \cite{michaelisBrian2LoihiEmulatorNeuromorphic2022} is not a deployment software for the Brian simulation framework, it is an emulator of the physical Loihi device.  So, it is not designed to run simulations on neuromorphic hardware.

Recent interoperability efforts such as the Neuromorphic Intermediate Representation (NIR) and SNN-MLIR address this broader fragmentation by defining common representations and compilation paths for spiking neural networks~\cite{pedersenNeuromorphicIntermediateRepresentation2024,generSNNMLIRMLIRDialect2026}. These efforts are complementary to NeuroML2Loihi, but the field still lacks a direct bridge from established neuroscience model-description standards such as NeuroML to Loihi-compatible primitives \cite{gleesonNeuroMLLanguageDescribing2010a,NextLevelNeuromorphicComputing}.

%% file: diagrams/fig_architecture.tex
\begin{figure*}[t]
  \centering
  \usetikzlibrary{arrows.meta}
  \colorlet{filedraw}{black!45}\colorlet{filefill}{black!5}
  \colorlet{stagedraw}{blue!55!black}\colorlet{stagefill}{blue!6}
  \colorlet{hidraw}{blue!62!black}\colorlet{hifill}{blue!16}
  \colorlet{backdraw}{teal!60!black}\colorlet{backfill}{teal!9}
  \colorlet{refdraw}{orange!72!black}\colorlet{reffill}{orange!12}
  \colorlet{edge}{black!55}
  \providecommand{\sub}{\footnotesize\color{black!72}}
  \resizebox{\textwidth}{!}{%
  \begin{tikzpicture}[
    font=\small,
    >={Latex[length=2.4mm]},
    every node/.style={outer sep=0pt},
    file/.style={draw=filedraw, thick, align=center, minimum height=0.95cm,
                 minimum width=2.4cm, fill=filefill, rounded corners=2pt},
    stage/.style={draw=stagedraw, thick, rounded corners=3pt, align=center,
                  minimum height=1.05cm, minimum width=2.6cm, fill=stagefill},
    hi/.style={stage, draw=hidraw, fill=hifill},
    feat/.style={draw=blue!72!black, very thick, rounded corners=3pt, align=center,
                 minimum height=1.55cm, minimum width=3.2cm, fill=blue!22},
    ext/.style={draw=teal!60!black, thick, dashed, rounded corners=3pt, align=center,
                minimum height=1.4cm, minimum width=3.4cm, fill=teal!9},
    lbl/.style={font=\footnotesize\color{black!70}},
    flow/.style={->, thick, draw=edge},
    extflow/.style={->, thick, dashed, draw=teal!55!black},
  ]
  \draw[draw=blue!45, very thick, dashed, rounded corners=6pt, fill=blue!2]
        (-1.5,4.25) rectangle (16.0,-2.15);
  \node[anchor=north west, font=\small\bfseries\color{blue!60!black}]
        at (-1.4,4.15) {NeuroML2Loihi framework};
  \node[file]  (nmlf) at (0,1.3)    {\textbf{NeuroML}\\[1pt]{\sub \texttt{.nml} model}};
  \node[file]  (lemf) at (0,-1.3)   {\textbf{LEMS}\\[1pt]{\sub \texttt{.xml} config}};
  \node[stage] (nmlp) at (3.3,1.3)  {\textbf{NMLParser}\\[1pt]{\sub $\rightarrow$ graph}};
  \node[stage] (lemp) at (3.3,-1.3) {\textbf{LEMSParser}\\[1pt]{\sub $\rightarrow$ config}};
  \node[stage] (net)  at (6.7,1.3)  {\textbf{Network}\\[1pt]{\sub NetworkX graph}};
  \node[stage] (cfg)  at (6.7,-1.3) {\textbf{Configuration}};
  \node[hi]    (map)  at (3.3,2.75) {\textbf{Mapping}\\[1pt]{\sub tags $\rightarrow$ Lava}};
  \node[stage] (lava) at (6.7,2.75) {\textbf{Lava primitives}\\[1pt]{\sub LIF / Dense}};
  \node[feat]  (exp)  at (10.8,0)   {\textbf{Experiment}\\[1pt]{\sub execute + record}};
  \node[stage] (ana)  at (14.4,0)   {\textbf{Analysis}\\[1pt]{\sub Van~Rossum,}\\{\sub Victor--Purpura}};
  \node[ext]   (back) at (10.8,-3.4)
    {\textbf{Backend}\\[1pt]{\sub Hardware: Loihi~2 (NxCore)}\\{\sub Simulator: LoihiSim}};
  \draw[flow] (nmlf) -- (nmlp);
  \draw[flow] (lemf) -- (lemp);
  \draw[flow] (nmlp) -- (net);
  \draw[flow] (lemp) -- (cfg);
  \draw[flow] (map)  -- (lava);
  \draw[flow] (lava) -- (net);
  \draw[flow] (cfg.north) -- node[right,lbl]{input nodes} (net.south);
  \draw[flow] (net.east) -- (exp.north west);
  \draw[flow] (cfg.east) -- (exp.south west);
  \draw[extflow] (back.north) -- node[right,lbl]{execution target} (exp.south);
  \draw[flow] (exp) -- (ana);
  \end{tikzpicture}%
  }
  \caption{The NeuroML2Loihi framework. A NeuroML model (\texttt{.nml}) and its LEMS
  simulation description (\texttt{.xml}) are read by separate parsers into a Network
  (a NetworkX graph) and a Configuration. Guided by the Mapping module, NeuroML components
  are assembled into Lava primitives (LIF/Dense) within the Network. The Experiment couples
  the Network and Configuration with an externally selected Backend---Loihi~2 hardware or the
  LoihiSim simulator---executes the run, and records outputs for Analysis.}
  \label{fig:architecture}
\end{figure*}
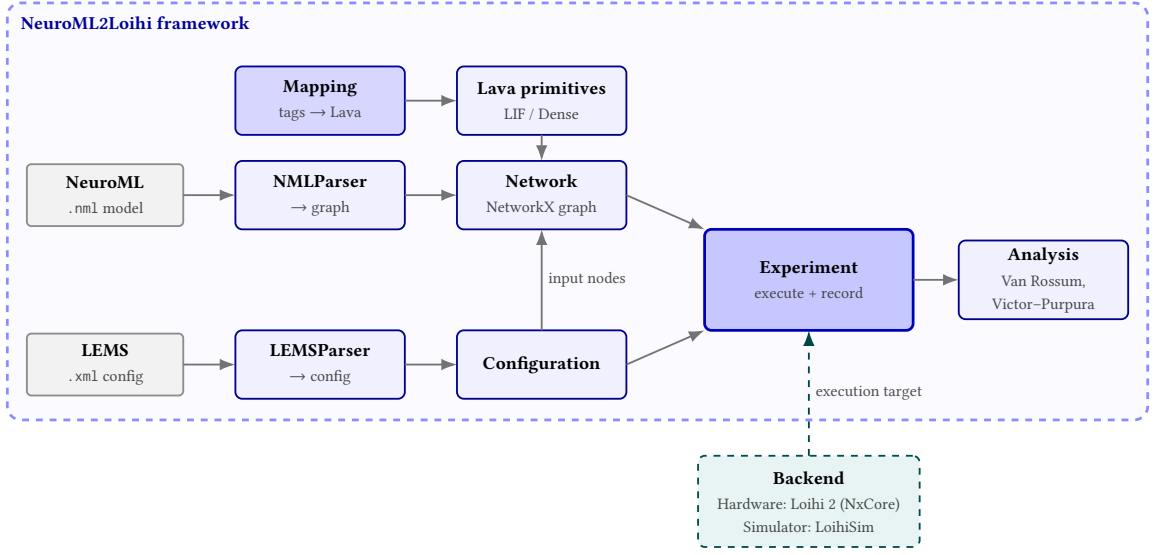

%% file: diagrams/fig1_lava_network.tex
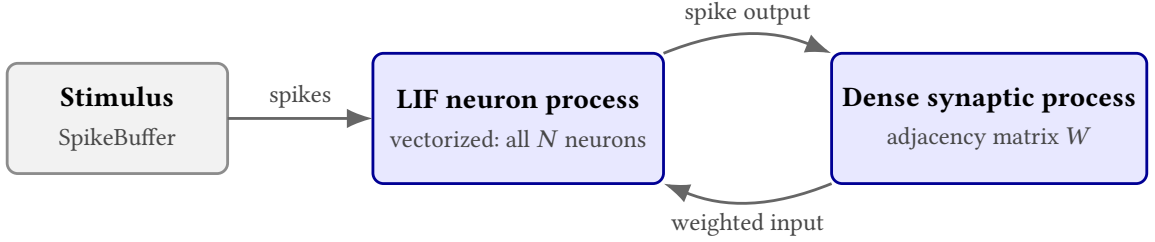
\begin{figure}[t]
  \centering
  \usetikzlibrary{arrows.meta}
  \colorlet{procdraw}{blue!58!black}\colorlet{procfill}{blue!9}
  \colorlet{iodraw}{black!45}\colorlet{iofill}{black!5}
  \colorlet{edge}{black!60}
  \providecommand{\sub}{\footnotesize\color{black!72}}
  \resizebox{\columnwidth}{!}{%
  \begin{tikzpicture}[
    font=\small,
    >={Latex[length=2.6mm]},
    every node/.style={outer sep=0pt},
    proc/.style={draw=procdraw, thick, rounded corners=3pt, align=center,
                 minimum height=1.3cm, minimum width=2.9cm, fill=procfill},
    io/.style={draw=iodraw, thick, rounded corners=3pt, align=center,
               minimum height=1.1cm, minimum width=2.2cm, fill=iofill},
    lbl/.style={font=\footnotesize\color{black!72}},
  ]
  \node[io]   (stim)  at (0,0)   {\textbf{Stimulus}\\[1pt]{\sub SpikeBuffer}};
  \node[proc] (lif)   at (4.0,0) {\textbf{LIF neuron process}\\[1pt]{\sub vectorized: all $N$ neurons}};
  \node[proc] (dense) at (8.7,0) {\textbf{Dense synaptic process}\\[1pt]{\sub adjacency matrix $W$}};
  \draw[->,thick,draw=edge] (stim) -- node[above,lbl]{spikes} (lif);
  \draw[->,thick,draw=edge] (lif.north east) to[bend left=22]
        node[above,lbl]{spike output} (dense.north west);
  \draw[->,thick,draw=edge] (dense.south west) to[bend left=22]
        node[below,lbl]{weighted input} (lif.south east);
  \end{tikzpicture}%
  }
  \caption{Vectorized Lava process network. All neurons are represented by a single LIF process
  connected recurrently to a Dense synaptic process through the adjacency matrix $W$ derived from
  the parsed NeuroML model.}
  \label{fig:lava-network}
\end{figure}

%% file: diagrams/fig2_hardware_pipeline.tex
\begin{figure*}[t]
  \centering
  \usetikzlibrary{arrows.meta}
  \colorlet{procdraw}{blue!58!black}\colorlet{procfill}{blue!9}
  \colorlet{iodraw}{black!45}\colorlet{iofill}{black!5}
  \colorlet{recdraw}{teal!60!black}\colorlet{recfill}{teal!9}
  \colorlet{edge}{black!58}
  \providecommand{\sub}{\footnotesize\color{black!72}}
  \resizebox{\textwidth}{!}{%
  \begin{tikzpicture}[
    font=\small,
    >={Latex[length=2.4mm]},
    every node/.style={outer sep=0pt},
    proc/.style={draw=procdraw, thick, rounded corners=3pt, align=center,
                 minimum height=1.15cm, minimum width=2.7cm, fill=procfill},
    io/.style={draw=iodraw, thick, rounded corners=3pt, align=center,
               minimum height=1.05cm, minimum width=2.4cm, fill=iofill},
    rec/.style={draw=recdraw, thick, rounded corners=3pt, align=center,
                minimum height=1.05cm, minimum width=2.5cm, fill=recfill},
    lbl/.style={font=\footnotesize\color{black!70}},
    flow/.style={->, thick, draw=edge},
  ]
  \draw[draw=teal!45, dashed, thick, rounded corners=6pt, fill=teal!3]
        (9.7,1.0) rectangle (20.3,-2.5);
  \node[anchor=south west, font=\footnotesize\bfseries\color{teal!55!black}]
        at (9.8,1.1) {hardware recording};
  \node[io]   (buf) at (0,0)   {\textbf{SpikeBuffer}\\[1pt]{\sub input pattern}};
  \node[proc] (din) at (3.7,0) {\textbf{Dense}\\[1pt]{\sub input weights}};
  \node[proc] (lif) at (7.4,0) {\textbf{LIF neuron process}\\[1pt]{\sub vectorized}};
  \draw[flow] (buf) -- node[above,lbl]{spikes} (din);
  \draw[flow] (din) -- (lif);
  \draw[flow] (lif.north east) to[out=125,in=55,looseness=0.85]
        node[midway,above,lbl]{recurrent Dense} (lif.north west);
  \node[proc] (drec) at (12.7,0)     {\textbf{Dense}\\[1pt]{\sub routing matrix}};
  \node[rec]  (recv) at (15.8,0)     {\textbf{Receiver}};
  \node[rec]  (sps)  at (18.9,0)     {\textbf{StateProbe}\\[1pt]{\sub spikes}};
  \node[rec]  (spv)  at (12.7,-1.75) {\textbf{StateProbe}\\[1pt]{\sub voltage ($v$)}};
  \draw[thick,draw=edge] (lif.east) -- ++(1.15,0) coordinate (J);
  \draw[flow] (J) -- node[above,lbl,pos=0.3]{spikes} (drec.west);
  \draw[flow] (J) |- node[pos=0.32,right=2pt,lbl]{$v$} (spv.west);
  \draw[flow] (drec) -- (recv);
  \draw[flow] (recv) -- (sps);
  \end{tikzpicture}%
  }
  \caption{Hardware input and recording pipeline. Input spikes drive the vectorized LIF process,
  which carries a recurrent Dense synapse. Spike outputs are routed through a Dense process to
  Receiver and StateProbe processes, while membrane voltage is recorded directly by a StateProbe
  on the neuron's $v$ property.}
  \label{fig:hw-pipeline}
\end{figure*}
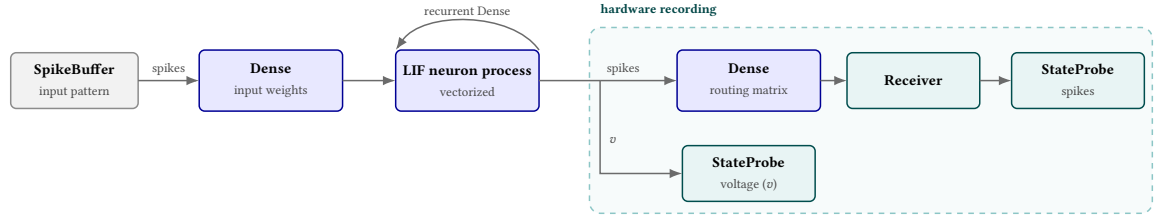

%% file: experiment_configurations.tex
\subsection{Experimental Configuration}
\label{sec:experimental-configuration}

This section describes the configuration for the experiments presented in this paper: the simulation hyperparameters, the neuron model and
its dynamics, the three connectome scales used as experimental networks, the
neurons selected for recording, and the two evaluation axes---energy
consumption and behavioral accuracy---together with the metrics used to
quantify them.

\subsubsection{Loihi 2 and Lava}
\label{subsubsec:loihi-2}
The neuromorphic hardware used for experimentation in our study is the Loihi-2 chip \cite{orchard2021efficient}, hosted on Intel's Kapoho Point Single Chip (KPSC) board. The chip features $128$ asynchronous ``Neuro-Cores'' that execute spiking networks. The chip also has $6$ low-power x86 processors called Lakemont (LMT) cores that  manage spike communication, data I/O, etc., and can execute 32-bit quantized non-spiking operations. \cite{gaurav2024legendre}. Each Neuro-Core supports $8192$ spiking units, thus, one Loihi-2 chip supports $1048576$ spiking neurons; although, the size of spiking networks deployed on Loihi-2 is determined not only by the number of neurons, but also the complexity of synaptic connections. The Loihi-2 chip supports learning mechanisms biology-inspired STDP and Reward-modulated STDP approaches.

Lava is Intel's open-source library for constructing, training, and deploying spiking networks on Loihi-2 neuromorphic chip \cite{lava}. The two core programming concepts in Lava are: \texttt{Process} and \texttt{ProcessModel}; each \texttt{Process} has its one or more corresponding \texttt{ProcessModel}s, depending on the hardware backend the \texttt{Process} is expected to be deployed. A \texttt{Process} represents the interface of the building blocks of a spiking network, e.g., a layer of LIF neurons, whereas, a \texttt{ProcessModel} implements the execution logic of the building block, e.g., the implementation of LIF neurons. Lava also provides a simulation backend of the Loihi-2 chip, which enables execution of spiking neural models on Loihi-2's simulation running on CPU; this assists greatly in quick prototyping and reproducibility.

\subsubsection{Neuron Model}
\label{subsec:neuron-model}
All neurons in our simulation are modeled as Leaky Integrate-and-Fire (LIF) units. The membrane
potential $v[t]$ evolves according to the standard LIF dynamics.  The \cd{Lava} implementation uses the following discrete-time LIF dynamics:

\begin{align}
u[t] &= (1 - d_u)u[t-1] + I[t], \label{eq:lava-lif-current} \\
v[t] &= (1 - d_v)v[t-1] + u[t] + b, \label{eq:lava-lif-voltage} \\
s[t] &= H\left(v[t] - v_{\mathrm{th}}\right), \label{eq:lava-lif-spike} \\
v[t] &\leftarrow 0 \quad \text{if } s[t] = 1, \label{eq:lava-lif-reset}
\end{align}
where $u[t]$ is the synaptic current state, $v[t]$ is the membrane voltage state,
$I[t]$ is the input at timestep $t$, $d_u$ and $d_v$ are the current and voltage
decay parameters, $b$ is a constant bias term, $v_{\mathrm{th}}$ is the firing
threshold, $H(\cdot)$ is the Heaviside step function, and $s[t]$ is the emitted
spike.  Rather than instantiating one \cd{Process} per neuron, all neurons at a given connectome scale are \emph{vectorized} into a single LIF process whose units are coupled through adjacency matrices representing the synaptic connectivity.

\subsubsection{Network models for Experimentation}
\label{subsec:networks}
To examine how energy and accuracy scale with network size, three subsets of the \emph{C.\ elegans} connectome are examined.  Muscle cells and neuron-to-muscle connections are excluded from the models to isolate stimuli to the PLM neurons (described below).

Synaptic connections are derived from keystone mappings of the full \textit{C. elegans} connectome by \cite{whiteStructureNervousSystem1986} and \citep{varshneyStructuralPropertiesCaenorhabditis2011}.  Connection weights are represented as integers and correspond to anatomical connection counts rather than measured synaptic strengths.  Because of the difference in scales between NEURON (\si{\milli\volt}) simulation and N2L simulation (Integer), input and recurrent weight scaling is used to enable matching behavior between simulation platforms.  Simulation parameters are modeled on Parameter Set A from \cite{gleeson2018c302}.

\paragraph{Model Set}
\begin{enumerate}
  \item \textbf{Full}: Includes the full set of 302 neurons in the \textit{C. elegans} connectome.  $3119$ synapses---excitatory, inhibitory, and gap junctions---are included.  
  \item \textbf{Active}: the subset of neurons that exhibit non-zero membrane voltage during full-scale Loihi simulation. The model includes 18 neurons and 125 synapses.  The synapse set is the subset of the full model synaptic set that exist between exclusively active neurons. Therefore, similar behavior between neurons in this model and neurons in the full model is expected.  This scale is used as a comparison to the full model to assess the energy consumption exclusively from the inactive neurons.
  \item \textbf{Minimal}: the PLML and PLMR neurons only, driven by direct input. No connections exist between these two neurons in this model.  This network is used to exclusively measure spiking activity.
\end{enumerate}
\begin{figure}[htbp]
    \centering
    \begin{minipage}{0.32\textwidth}
        \centering
        \includegraphics[width=\textwidth]{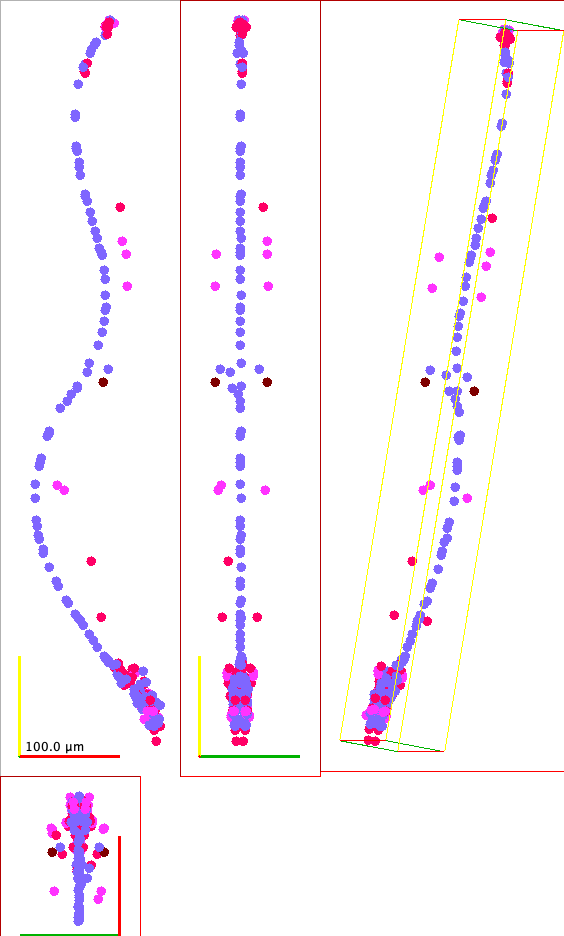}
        \textbf{A}
    \end{minipage}
    \hspace{70px}
    \begin{minipage}{0.32\textwidth}
        \centering
        \includegraphics[width=\textwidth]{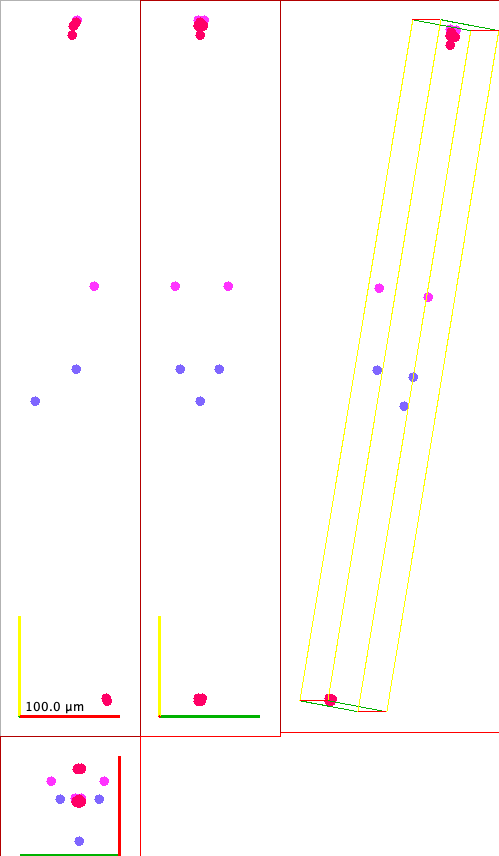}
        \textbf{B}
    \end{minipage}

    \caption{Visualization of the Full (\textbf{A}) and Active (\textbf{B}) neuron subsets in the c302-derived NeuroML2 networks used in this study. Figures were generated using from the corresponding NeuroML model files used for simulation using pyNeuroML~\cite{vellaLibNeuroMLPyLEMSUsing2014}.}
    \label{fig:c302-active-neurons}
\end{figure}
\paragraph{Neuron Stimulation (Inputs)}
In all models, the posterior lateral microtubule neurons PLML and PLMR are stimulated. These neurons are located in the posterior region of the C. elegans and function as posterior touch receptor neurons \cite{chalfieNeuralCircuitTouch1985}. Their stimulation is therefore useful for probing the neural response to posterior touch. By examining the activation of locomotion-related neurons following PLM stimulation, the corresponding motor response can be inferred from the pattern of downstream neural activity. This behavior is examined in further details in the Section \ref{sec:discussion}  Discussion.  During experimentation, stimulation begins at timestep 0 and persists throughout the duration of the experiment (\qty{300}{\milli\second}).   

\subsubsection{Hyperparameters}
\label{subsec:hyperparameters}

Each experiment is run for \qty{300}{\milli\second} of biological time with a timestep length of of \qty{0.02}{\milli\second} of biological time yielding $15,000$ timesteps total. These values are chosen to capture the transient and early steady-state behavior of the neurons while remaining tractable under the limited resources of a single Loihi~2 chip.

The set of hyperparameters used during our experimentation is shown in Table~\ref{tab:model-parameters}.  
\begin{table}[ht]
    \centering
    \caption{Model parameters.}
    \label{tab:model-parameters}
    \begin{tabular}{lc}
        \hline
        \textbf{Parameter} & \textbf{Value} \\
        \hline
        $d_u$ & $49$ \\
        $d_v$ & $8$ \\
        $v_{\mathrm{th}}$ & $131,071$ \\
        Recurrent scale & $10\cdot 2^5=320$ \\
        Input scale & $4$ \\
        Biological duration & \qty{300}{\milli\second} \\
        Timestep length (Biological) & \qty{0.02}{\milli\second} \\
        
        \hline
    \end{tabular}
\end{table}
Loihi threshold is set to $131,071$, the maximum allowed by Loihi to capture maximal granularity in subthreshold neural dynamics. Additionally, in order to capture the fluctuating currents represented in NeuroML \texttt{expTwoSynapse} synapse mode, the \textit{current} and \textit{voltage} decay attributes of the Lava \cd{LIF} neuron ($du$ and $dv$ respectively) are leveraged.  To optimize accuracy of benchmarking results, these values were determined through grid search hyperparameter tuning methods.

%% file: benchmarking_methods.tex
\subsection{Benchmarking}
\input{accuracy_methods}

\input{latency_methods}

\input{energy_methods}

\input{memory_methods}

%% file: accuracy_methods.tex
  \subsubsection{Cross-platform fidelity metrics}
  \label{sec:compare-metrics}

  We quantify how faithfully the Loihi~2 hardware reproduces the reference
  OpenWorm simulation by comparing their spike trains and membrane
  potentials neuron-by-neuron, aligned by neuron name. For neuron $i$ we write its spike train as the set of spike times $X_i=\{t_{i,1},t_{i,2},\dots\}$ and
  its membrane potential as $v_i(t)$; metrics are computed per neuron and
  averaged over the set of neurons active on at least one platform.

  \paragraph{Victor--Purpura distance.}
  $D_{\mathrm{VP}}^{q}(X,Y)$ is the minimum total cost to transform spike train $X$
  into spike train $Y$ using three elementary operations: inserting a spike (cost
  $1$), deleting a spike (cost $1$), and shifting a spike in time by $\Delta t$
  (cost $q\,\lvert\Delta t\rvert$)
   \cite{victor1996nature}. The cost parameter $q$
  (units of $\mathrm{s}^{-1}$) sets the timescale at which timing is penalized:
  at $q=0$ the distance reduces to the spike-count difference, while larger $q$
  increasingly penalizes temporal misalignment. We report
  $q\in\{1,50\}\,\mathrm{s}^{-1}$. $q=1$ is used such that these results can be comparable to other experiments. $q=50$ was selected to approximate the reference simulations' PLM neurons' Inter-Spike-Interval (ISI).  At a $~63$~Hz spike rate, the ISI of PLM neurons in the OpenWorm simulation is around $16$ms.  $q=50$ yields a shift penalty of $20$ ms.

  \paragraph{van~Rossum distance.}
  Each spike train is convolved with a causal exponential kernel
  $k(t)=e^{-t/\tau}\,\mathbb{1}[t\ge 0]$ to yield
  $f_X(t)=\sum_{t_j\in X}k(t-t_j)$, and the distance is the normalized $L_2$
  difference of the filtered signals~\cite{van2001novel},
  \begin{equation}
    D_{\mathrm{vR}}^{\tau}(X,Y)=
    \sqrt{\frac{1}{\tau}\int_{0}^{\infty}\!\bigl(f_X(t)-f_Y(t)\bigr)^2\,dt}\,,
  \end{equation}
  with time constant $\tau=20\ \mathrm{ms}$ as calculated per the convention of $\tau=\frac{1}{q}$.

  \paragraph{Membrane-potential agreement.}
  Because the platforms use different voltage scales, the raw Loihi~2 trace
  $v^{\mathrm{HW}}$ is mapped onto the reference $\mathrm{mV}$ scale by the affine
  transform anchored on the resting and threshold potentials,
  \begin{equation}
    \tilde v = g_{\mathrm{rest}}
    + \bigl(v^{\mathrm{HW}}-h_{\mathrm{rest}}\bigr)\,
    \frac{g_{\theta}-g_{\mathrm{rest}}}{h_{\theta}-h_{\mathrm{rest}}},
  \end{equation}
  where $(h_{\mathrm{rest}},h_{\theta})$ and $(g_{\mathrm{rest}},g_{\theta})$ are
  the resting and threshold values on hardware and reference, respectively.
  Per-neuron agreement is reported as the Pearson correlation $\rho$, where:
  \begin{equation}
    \rho=
    \frac{\sum_{m}\bigl(\tilde v_m-\bar{\tilde
  v}\bigr)\bigl(v^{\mathrm{ref}}_m-\bar{v}^{\mathrm{ref}}\bigr)}
         {\sqrt{\sum_{m}\bigl(\tilde v_m-\bar{\tilde v}\bigr)^2}\,
          \sqrt{\sum_{m}\bigl(v^{\mathrm{ref}}_m-\bar{v}^{\mathrm{ref}}\bigr)^2}}
  \end{equation}
  (capturing trace \emph{shape}, scale-invariant) together with the magnitude
  errors over the $M$ shared time samples,
  \begin{equation}
    \mathrm{RMSE}=\sqrt{\frac{1}{M}\sum_{m=1}^{M}\bigl(\tilde v_m-v^{\mathrm{ref}}_m\bigr)^2},
    \qquad
    \mathrm{MAE}=\frac{1}{M}\sum_{m=1}^{M}\bigl\lvert\tilde v_m-v^{\mathrm{ref}}_m\bigr\rvert,
  \end{equation}
  both expressed in $\mathrm{mV}$.

  \paragraph{Poisson baseline.}
  To calibrate these distances against chance, each spike train is also scored
  ($D_{\mathrm{VP}}^{q}$, $D_{\mathrm{VR}}^{\tau}$) against a homogeneous Poisson
  reference train of matched mean rate $\lambda$ (exponentially distributed ISI
 $\lambda=63\,\mathrm{Hz}$). We report the ratio
  $D_{\text{Loihi~2 vs.\ ref}}/D_{\text{Loihi~2 vs.\ Poisson}}$; a value below $1$
  indicates that Loihi~2 is closer to the reference simulation than to a
  rate-matched random process.

%% file: latency_methods.tex
\subsubsection{Latency Benchmarking}
\label{subsec:latency}

The wall-clock cost of each run is measured on a cohort of $25$ seeds and three connectome scales. On Loihi~2, latency is taken as the on-chip execution
window measured by the power profiler which is the closest available proxy for pure inference latency. Computed time-steps' latency is then normalized by the biological duration of the run to give a Real-Time Factor (RTF),
\begin{equation}
  \mathrm{RTF} = \frac{T_{\mathrm{wall}}}{T_{\mathrm{bio}}},
  \label{eq:rtf}
\end{equation} the number of wall-clock seconds spent per simulated second; an
$\mathrm{RTF} < 1$ indicates faster-than-biological real time. Per-model figures
are aggregated across seeds into a mean, standard deviation, and $95\%$
confidence interval, and the platforms are compared through a speedup ratio.

%% file: energy_methods.tex
\subsubsection{Energy Measurement}
\label{subsubsec:energy}

On Loihi~2, energy is measured by sampling instantaneous power throughout 
execution and integrating it over time, rather than reading a single aggregate counter. This yields a power trace per run from which both energy and mean power are derived.  A \texttt{Loihi2Power} profiler is registered as a run callback so
that the on-chip power rails are sampled while the network executes. Each
sample records the total board power and its static component, together with
the per-rail breakdown ($V_{\mathrm{dd}}$, $V_{\mathrm{ddm}}$,
$V_{\mathrm{ddio}}$) and a microsecond timestamp. Energy is obtained by trapezoidal integration of each power series against time,
\begin{equation}
  E = \int_{t_0}^{t_1} P(t)\,\mathrm{d}t,
  \label{eq:energy-integral}
\end{equation} and mean power is calculated as $\bar{P} = E / (t_1 - t_0)$. The total chip energy ($E_{\mathrm{total}}$, the integral of total board power over the execution
window). 

%% file: memory_methods.tex
\subsection{Memory Collection}
Loihi 2 memory occupancy was measured using the \texttt{Loihi2Memory} profiler. Each Loihi 2 neurocore provides $192\,\mathrm{KB}$ of local memory, which can be flexibly allocated across neuron state, synaptic state, routing, and other mapped structures~\cite{NextLevelNeuromorphicComputing}. Internally this memory is partitioned into a small-entry and a large-entry pool that occupy $1/5$ and $4/5$ of the core respectively, and the profiler reports the occupied ratio of each pool for every mapped core. The total occupied ratio of core $c$ is their capacity-weighted sum,

\begin{equation}
r_c = \tfrac{1}{5}\,r_c^{\mathrm{small}} + \tfrac{4}{5}\,r_c^{\mathrm{large}},
\end{equation}

so that $r_c$ reflects the fraction of the core's $192\,\mathrm{KB}$ that is in use. Total memory usage is then

\begin{equation}
M_{\mathrm{Loihi}} = \sum_{c \in C} r_c \cdot 192\,\mathrm{KB},
\end{equation}

where $C$ is the set of mapped neurocores. The result is converted to megabytes and reported as the Loihi's SRAM footprint of the deployed model. Because the evaluated networks are statically compiled before execution, mapped memory occupancy is fixed for each run.

%% file: openworm_simulation_comparison.tex
\subsection{NEURON Reference Simulation}
\label{subsec:openworm-config}

Experiments using the same models as those examined for Loihi~2 are executed using the NEURON simulator on GPU-based hardware to serve as a reference baseline to convention computing platforms. The reference uses an identical experimental design---OpenWorm \emph{parameter set A}, the same three connectome scales
(Section~\ref{subsec:networks}), and the same cohort of random seeds---so that
each reference run pairs one-to-one with a hardware run and the two differ only
in their execution substrate. This subsection describes that reference
configuration and the measurements collected from it; the corresponding
hardware procedures are given in Sections~\ref{subsubsec:ow-energy}
and~\ref{subsubsec:ow-latency}.  Memory is not used for comparison for the NEURON simulation because Linux process memory metrics (e.g., resident set size) are not directly comparable to Loihi~2’s neurocore occupancy metric.  However, memory metrics for the NEURON reference simulations can be found in Supplementary tables S17--S19.

\subsubsection{Simulator and Platform}
\label{subsubsec:ow-simulator}

The reference networks are run on the NEURON
simulator~\cite{hinesNEURONSimulationEnvironment1997}. Execution is delegated to the CoreNEURON
compute engine~\cite{kumbharCoreNEURONOptimizedCompute2019}, which compiles the model to a GPU
backend; the simulations are run on an NVIDIA Jetson AGX Orin at 15W mode. Note that Jetson AGX Orin is an edge computing device specifically designed for low power consumption.  

Each neuron is integrated with a fixed timestep of $\Delta t = 0.02$~ms over
the same $300$~ms biological window as the Loihi~2 runs, matching the model's
dynamics rather than the event-driven execution of Loihi~2. Only the neural
model is simulated: the Sibernetic body--environment physics of the full
OpenWorm stack is disabled, so the reference measures the cost of the neuronal
simulation alone and is comparable to the network deployed on Loihi~2.

\subsubsection{Energy Measurement}
\label{subsubsec:ow-energy}

Energy on the reference platform is measured by sampling the Jetson's
instrumented power rails throughout execution. Separate CPU and GPU rails are read at  $49\text{--}50$~Hz---on the Jetson from its integrated INA3221 power monitors.  Each
rail's power trace is integrated over time to yield energy, exactly as on
Loihi~2 (Eq.~\ref{eq:energy-integral}).  Crucially, the reported reference energy is \emph{dynamic}: a \qty{30}{\second} idle baseline is recorded immediately before each simulation, and the mean idle power of each rail is subtracted from the trace so that only the compute-attributable, activity-driven energy remains. The CPU and GPU dynamic energies are then summed into a single system-compute figure. This idle-relative definition is deliberately distinct from the Loihi~2 measurement, which integrates \emph{total} chip power inclusive of static draw (Section~\ref{subsubsec:energy}); the consequence of that asymmetry is discussed in Section~\ref{subsubsec:asymmetric-energy}.

\subsubsection{Latency Measurement}
\label{subsubsec:ow-latency}

Latency on the reference platform is taken as the simulation-loop (compute)
time---the wall-clock duration of the integration over the $300$~ms window,
excluding the one-off network generation and model compilation, which on this
toolchain dominate setup, but are amortized once per model. This is the fair
counterpart to the on-chip execution window measured on Loihi~2
(Section~\ref{subsec:latency}), and the two are compared through the
duration-normalized Real-Time Factor of Eq.~\ref{eq:rtf}.

\subsubsection{Asymmetric Energy Comparison}
\label{subsubsec:asymmetric-energy}

The two platforms are therefore credited differently intentionally:\textbf{ the reference figure is the host's \emph{dynamic} (idle-subtracted) compute energy,
whereas the Loihi~2 figure is the chip's \emph{total} energy including its static draw}. The comparison is thus conservative \textit{toward} the neuromorphic hardware.  So, any energy advantage reported for Loihi~2 can be considered a lower bound on the true gap.  Total energy and wattage without a subtracted idle baseline can be found in Supplemental Tables S8--S10.

%% file: similarity_results.tex
\subsection{Similarity}
\begin{table}[t]
  \centering
  \caption{NeuroML2Loihi vs NEURON}
  \label{tab:n2l-vs-ow-spikes}
  \begin{tabular}{lrrr}
    \toprule
    metric & N2L vs NEURON & N2L vs Poisson & Ratio \\
    \midrule
    VP@1 & 1.0914 & 3.8133 & 0.2862 \\
    VP@50 & 5.5700 & 12.2079 & 0.4563 \\
    VR & 3.4194 & 4.7154 & 0.7252 \\
    $\lvert \Delta \mathrm{spikes}\rvert /\mathrm{neuron}$ & 1.0000 & 3.5000 & 0.2857 \\
    \bottomrule
  \end{tabular}
\end{table}

\begin{table}[t]
  \centering
  \caption{Voltage (NeuroML2Loihi affine-mapped to NEURON voltage scales)}
  \label{tab:n2l-vs-ow-voltage}
  \begin{tabular}{lrrr}
    \toprule
    metric & PLM & Active & Full \\
    \midrule
    Active Neurons Loihi~2 & 2 & 18 & 18 \\
    Active Neurons NEURON & 2 & 13 & 13 \\
    voltage r & -0.1441 & 0.4334 & 0.4334 \\
    voltage RMSE ($mV$) & 8.8322 & 3.9428 & 0.2350 \\
    voltage MAE ($mV$) & 7.2744 & 3.3601 & 0.2003 \\
    \bottomrule
  \end{tabular}
\end{table}
The results of similarity experimentation and comparison are represented in Tables \ref{tab:n2l-vs-ow-spikes} and \ref{tab:n2l-vs-ow-voltage}.  Due to the deterministic behavior of the neuron dynamics across experiments, the standard deviation across all seeds is $0$ for these particular metrics.  Additionally, action potentials were limited to PLM neurons across all simulations.  So, comparison results are identical across all models.  Thus, spiking comparison results have been consolidated into a single Table (\ref{tab:n2l-vs-ow-spikes}).

\subsubsection{Spiking Results}

Overall, spike train comparison depicts faithful spike timing replication on N2L simulation compared to NEURON simulation.  Victor--Purpura ratio shows the N2L simulator is less than half as expensive ($45\%$) to transfer to the NEURON simulator results when $q=50$.  This can be interpreted as the N2L simulation produces spike trains that are twice as close to the NEURON reference spike trains as the rate-matched baseline.  Van Rossom (VR) distance relative to baseline also shows a distance reduction with ratio $\approx73\%$.  

\subsubsection{Voltage Results}
Table~\ref{tab:n2l-vs-ow-voltage} reports voltage-trace similarity after affine mapping the NeuroML2Loihi traces onto the OpenWorm voltage scale. In contrast to the spike-train metrics, the voltage results show weaker cross-platform agreement, indicating that NeuroML2Loihi under these experimental configurations more faithfully preserves spike timing than it preserves detailed subthreshold membrane dynamics.

\begin{figure}[t]
    \centering
    \includegraphics[width=0.85\linewidth]{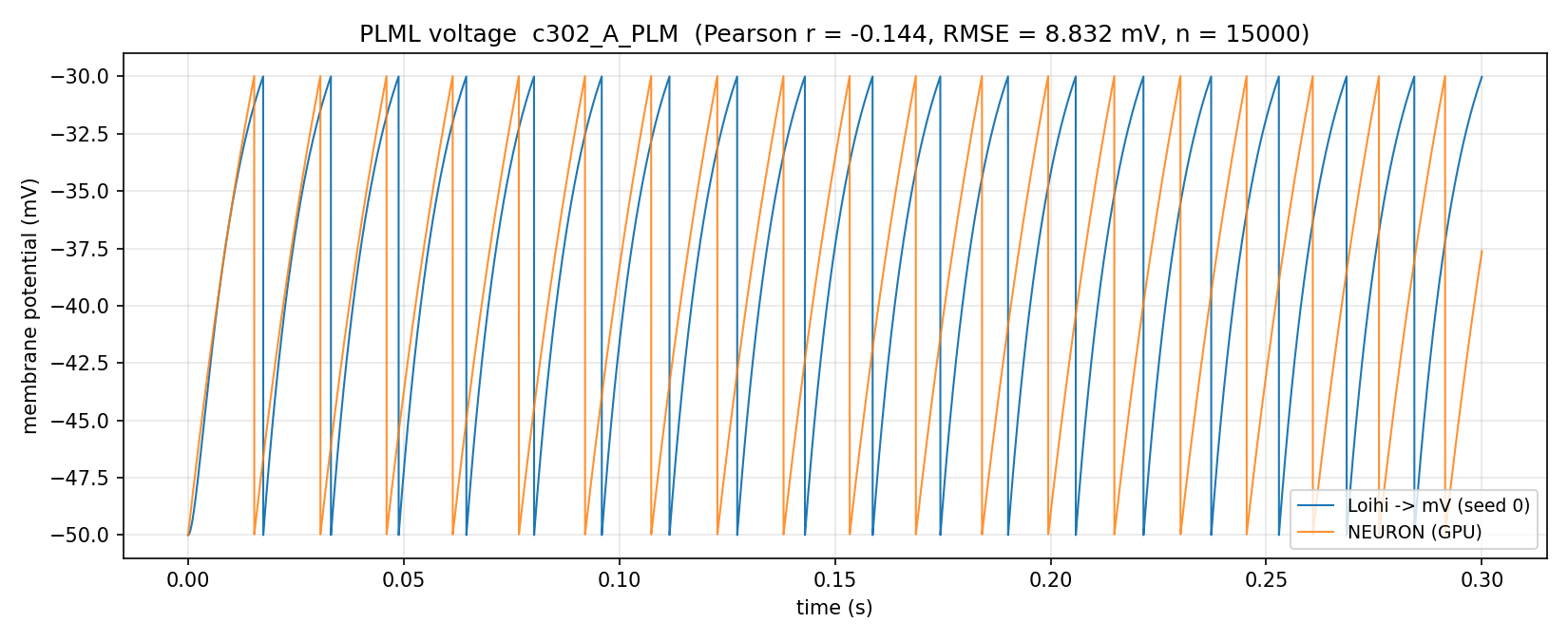}
    \caption{Superimposed PLML voltage traces from the NeuroML2Loihi and NEURON simulations. The comparison illustrates that the platforms produce similar structured voltage evolution while differing in the exact subthreshold trajectory due to the use of Lava LIF and Dense primitives on Loihi versus the NeuroML/c302 neuron and synapse dynamics used by the reference simulator.}
    \label{fig:plml-voltage-overlay}
\end{figure}

The minimal PLM-only model shows the weakest voltage agreement, with a negative Pearson correlation ($r=-0.1441$). This appears to result from a consistent offset in the oscillatory PLM voltage traces: peaks and troughs occur at slightly different times across platforms, causing the traces to appear poorly correlated despite exhibiting similar structured fluctuations.  Such differentiation can be seen clearly in Figure \ref{fig:plml-voltage-overlay}. 

The active-neuron and full models show moderate Pearson agreement ($r=0.4334$). The full model has substantially lower RMSE and MAE, but this is partly due to the large number of neurons that remain near quiescent voltage levels. Therefore, the reduced absolute error in the full model should not be interpreted as uniformly stronger voltage fidelity across all neurons.  Representative voltage trace comparisons for all active recorded neurons are provided in Supplementary Figures S1–S6.

Overall, the voltage mismatch is expected given the different simulation primitives used by each platform. The NEURON reference uses \cd{IAF} (Integrate and Fire) neurons and \cd{ExpTwoSynapse} models, while NeuroML2Loihi uses Lava \texttt{LIF} neurons and \texttt{Dense} synaptic processes. These implementations can reproduce similar spike-level behavior with properly tuned neuron-parameters while still producing non-identical membrane voltage trajectories.


Note that matching behavior across substrates requires tuning N2L experiment hyperparameters: input scaling, recurrent scaling, and decay terms, which makes a comparison experiment necessary whenever a new NeuroML model is deployed through NeuroML2Loihi. In principle these parameters could be derived directly from the NeuroML and NEURON configuration to give a more automatic mapping; developing that derivation is left to future work.

Thus, our results indicate that biological faithfulness is attainable on Loihi~2 for this class of small-scale simulation, particularly in the spiking regime. The mismatch in voltage regime is not a fundamental hardware limitation but a consequence of the software-level model mapping: Lava \texttt{LIF} neurons with \texttt{Dense} connections are an accessible, hardware-compatible target but only approximate the reference primitives. The current implementation therefore establishes a conservative baseline in which NeuroML2Loihi reproduces the primary spike-level behavior while preserving enough network behavior to support the biological interpretation.

%% file: benchmarking_metrics_results.tex
\subsection{Benchmarking Metrics}
\begin{table}[htbp]
  \centering
    \setlength{\tabcolsep}{3pt} 
    \caption{N2L on Loihi~2 vs NEURON on Jetson -- Latency and Energy}
  \label{tab:n2l-vs-neuron-main-comparison}
  \begin{tabular}{@{}lr|r|r|r|r|r}
    \toprule
    Model & N2L latency (s) & NEURON run time (s) & Speedup & N2L energy (J) & NEURON energy (J) & Energy (N2L / NEURON) \\
    \midrule
    Minimal & $0.0243 \pm 0.0001$ & $12.47 \pm 0.11$ & $514\times $ & $0.0122 \pm 0.0001$ & $11.65 \pm 0.89$ & 0.105\% \\
    Active & $0.0242 \pm 0.0001$ & $18.62 \pm 0.09$ & $768\times $ & $0.0123 \pm 0.0001$ & $18.11 \pm 0.36$ & 0.0676\% \\
    Full & $0.0258 \pm 0.0001$ & $31.19 \pm 0.12$ & $1211\times $ & $0.0131 \pm 0.0001$ & $28.52 \pm 0.55$ & 0.0458\% \\
    \bottomrule
  \end{tabular}
\end{table}

Table \ref{tab:n2l-vs-neuron-main-comparison} depicts an overview of metrics comparisons for Latency and Energy consumption of C. elegans neural simulation on different hardware backends.  Units are normalized across each category to enable easy comparison and $95\%$ Confidence Interval is also included.  Overall, N2L simulation on Loihi~2  achieves significant advantages across each category.  Per-seed energy metrics for N2L on Loihi~2 and NEURON on Jetson can be found in Supplementary Tables S2--S7.  Per-seed latency results can be found in Supplementary Tables S11--S16.

Across all three model scales, NeuroML2Loihi execution on Loihi 2 substantially reduced latency and energy consumption relative to the NEURON reference simulation on Jetson. The largest effect was observed in latency and energy. Loihi~2 execution remained nearly constant across model scale, increasing only slightly from the Minimal to Full connectome ($<6\%$), whereas NEURON runtime and energy increased monotonically as network size increased. Consequently, the relative advantage of Loihi~2 increased with scale: speedup rose from 514× on the Minimal model to 1211× on the Full model, while Loihi energy fell from $0.105\%$ to $0.0458\%$ of the NEURON energy.

These results indicate that the neuromorphic implementation does not merely provide a fixed-factor improvement. Instead, the advantage grows with model scale because the Loihi~2 execution cost is dominated by a nearly constant on-chip execution window over this range, while the conventional simulation cost increases with model size.

\begin{table}[!ht]
  \centering
  \caption{Power comparison between N2L simulation on Loihi-2 vs NEURON on Jetson}
  \label{tab:n2l-vs-ow-power}
  \begin{tabular}{lr|r|r|r}
    \toprule
    Model & N2L power (W) & NEURON power (W) & Power ratio (NEURON / N2L) \\
    \midrule
    Minimal &  $0.5047 \pm 0.0018$ & $0.9393 \pm 0.0706$ & 1.9x \\
    Active &  $0.5052 \pm 0.0020$ & $0.9778 \pm 0.0199$ & 1.9x \\
    Full & $0.5067 \pm 0.0020$ & $0.9189 \pm 0.0178$ & 1.8x \\
    \bottomrule
  \end{tabular}
\end{table}

Table~\ref{tab:n2l-vs-ow-power} reports mean power draw during simulation execution. In contrast to latency and energy, power varies only modestly across model scale on both platforms. Loihi power remains nearly constant across the Minimal, Active, and Full models, ranging from $0.5047$ to $0.5067$ W. NEURON on Jetson power is also relatively stable, ranging from $0.9189$ to $0.9778$ W, with overlapping $95\%$ confidence intervals across models. As a result, the instantaneous power advantage of Loihi remains approximately constant, with NEURON on Jetson drawing only $1.8$–$1.9\times$ more power than Loihi.

This exposes an important distinction between power and energy in the benchmark results. Although Loihi does draw less power during execution, the much larger energy reduction in Table~\ref{tab:n2l-vs-neuron-main-comparison} is primarily explained by the reduction in simulation time rather than by a proportionally large reduction in instantaneous power. Thus, the energy advantage arises from the combination of a modest power reduction and a substantially larger latency reduction.

\begin{table}[htbp]
  \centering
  \caption{Loihi on-chip memory footprint}
  \label{tab:memory-summary-loihi-on-chip-memory-footprint}
  \begin{tabular}{lrrr}
    \toprule
    Model & Cores & Footprint (MB) & Total \% mean \\
    \midrule
    Minimal & 3 & 0.001686 & 0.29966 \\
    Active & 3 & 0.005424 & 0.96418 \\
    Full & 9 & 0.6571 & 38.93860 \\
    \bottomrule
  \end{tabular}
\end{table}

Table~\ref{tab:memory-summary-loihi-on-chip-memory-footprint} reports the Loihi 2 on-chip memory footprint of each deployed model. The Minimal and Active models occupy only a small fraction of the available mapped core memory, using $0.001686\,\mathrm{MB}$ and $0.005424\,\mathrm{MB}$ across three cores, respectively. The Full model requires nine cores and occupies $0.6571\,\mathrm{MB}$, corresponding to a mean total occupancy of $38.94\%$. Thus, all evaluated networks fit comfortably on a single Loihi~2 device, but memory usage increases disproportionately with the transition from the Active to Full model. The scaling implications of this increase in mapped memory usage are discussed in Section~\ref{subsubsec:memory-implications}.

%% file: benchmarking_analysis_discussion.tex
\subsection{Benchmarking Analysis}
\subsubsection{Energy Advantage}
The benchmarking results demonstrate that NeuroML2Loihi enables low-energy neural simulation on Loihi~2 while preserving the functional behavior discussed in the preceding similarity analysis. Across the three evaluated connectome scales, Loihi~2 execution produced substantially lower energy consumption than the NEURON reference simulation. This advantage was not primarily the result of a large reduction in instantaneous power draw. Rather, the dominant factor was the much shorter execution duration enabled by the neuromorphic substrate.

This distinction is important because power and energy capture different aspects of the system behavior. Mean power draw on Loihi~2 was lower than the NEURON reference on Jetson, but only by a factor of approximately 1.8--1.9$\times$. In contrast, total energy consumption differed by orders of magnitude. Therefore, the observed energy advantage is best understood as the compound effect of modestly lower power and dramatically reduced latency. Loihi~2 does not merely consume less power while performing the same computation over the same interval; it completes the neural simulation in a much shorter wall-clock time, reducing the duration over which power is drawn.

This behavior is consistent with the architectural principles of neuromorphic computing. Loihi~2 implements neural computation using event-driven execution and localized on-chip state, reducing the dependence on repeated movement of neural and synaptic state through a conventional memory hierarchy. For neural simulation workloads, this is especially valuable because neurons and synapses are repeatedly updated over many timesteps. When the relevant state can remain close to the compute substrate and computation is driven by neural events, the simulation can avoid much of the memory movement and control overhead associated with conventional step-based execution. The result is a substantial reduction in latency, which in turn produces the observed reduction in total energy.

The low Loihi~2 energy values should therefore be interpreted as evidence that the neuromorphic implementation is well matched to the structure of the simulated workload. The \textit{C.~elegans} networks evaluated here are sparse, recurrent neural systems with localized communication patterns and limited spiking activity. These properties align naturally with the design goals of neuromorphic hardware: representing neural state directly on chip, communicating through spikes, and advancing computation through event-driven mechanisms.

Additionally, latency is itself a central metric for neural simulation. Faster simulation enables larger experimental sweeps, more random seeds, longer biological durations, and more extensive hyperparameter exploration under the same wall-clock budget. In this context, latency reduction is not only a performance improvement but also an enabler of scientific productivity. A simulator that can complete equivalent biological experiments more quickly allows researchers to evaluate more model variants and perform more rigorous validation within practical time constraints.

These benchmarking results indicate that NeuroML2Loihi leverages Loihi~2's neuromorphic architecture to provide substantial energy advantages for this class of neural simulation workload. The results should not be interpreted as showing that Loihi~2 has a proportionally large instantaneous power advantage. Instead, they show that brain-inspired architectures prompt reduced execution time which can translate into orders-of-magnitude lower energy consumption for NeuroML-derived neural simulations.

\subsubsection{Memory Occupancy Implications}
\label{subsubsec:memory-implications}
The Loihi mapping results show that scalability cannot be discussed \textit{only} in terms of neuron count. Between the Active and Full models, the number of neurons increases from 18 to 302, a factor of approximately 16.8. Over the same transition, mapped Loihi SRAM occupancy increases from 0.005424MB to 0.6571MB, a factor of approximately 121. This disproportionate increase indicates that the hardware mapping cost is driven not only by the number of neurons, but also by the recurrent synaptic structure required to connect them. This is expected for biologically derived networks, where synaptic connectivity can dominate the memory and routing cost of deployment.

This observation is especially important when considering larger nervous systems. Biological neurons may form thousands or tens of thousands of synaptic connections, and therefore a model with more neurons is not merely a larger collection of independent units. It is a larger communication and state-management problem. For neuromorphic simulation, the limiting factor may therefore become the representation, routing, and storage of synaptic connectivity rather than the raw number of neuron compartments alone. The \textit{C. elegans} Full model remains small enough to fit comfortably on a single Loihi~2 device, but the Active-to-Full scaling behavior suggests that synaptic density must be treated as a key scalability constraint in future work.

%% file: conclusion.tex
\section{Conclusion and Future Work \label{sec:conc-fw}}

This work introduced NeuroML2Loihi, an open-source framework for translating NeuroML neural model descriptions into Loihi~2-deployable Lava primitives. 
NeuroML is already used by the computational neuroscience community to define neurons, synapses, networks, and simulation configurations. By translating NeuroML-derived models into Loihi~2-native primitives, NeuroML2Loihi provides a bridge between an existing modeling standard and the energy advantages of neuromorphic execution. This design choice is central to scalability: a translation framework built around a community modeling standard allows neuromorphic experimentation to be applied to models that were not originally designed for neuromorphic deployment. Thus, the NeuroML2Loihi framework addresses a gap between established computational neuroscience tooling and neuromorphic hardware workflows by allowing standardized NeuroML models to be executed on Loihi 2 while preserving comparison against conventional  outputs. Using c302-derived \textit{C. elegans} models, NeuroML2Loihi was evaluated across Minimal, Active, and Full connectome-scale networks. The Loihi~2 implementation reproduced the primary spike-level behavior of the OpenWorm reference simulation, although voltage-trace agreement was weaker due to differences between the Lava LIF/Dense mapping and the original NeuroML/c302 neuron and synapse dynamics. Thus, the present implementation demonstrates functional spike-level fidelity rather than exact subthreshold equivalence. Benchmarking showed substantial latency and energy advantages for Loihi~2 on the evaluated workloads. The energy reduction was driven primarily by dramatically shorter execution time, combined with lower instantaneous power. The results also show that scaling is not determined by neuron count alone: recurrent synaptic connectivity substantially affects Loihi~2 memory occupancy and should be treated as a central constraint in larger deployments. Future work should extend NeuroML2Loihi to larger and more densely connected NeuroML models, longer biological simulation durations, and more complex neuron and synapse dynamics  -- not only on a single Loihi-2 chip, but also on neuromorphic boards/systems comprising multiple Loihi-2 chips. Another direction could be to improve voltage-level agreement through custom Lava neuron or synapse models that more closely match the NeuroML primitives, or through custom NeuroML components designed for Loihi~2 deployment. A systematic method for deriving Loihi~2-side parameters from NeuroML and LEMS configurations would also reduce the need for manual hyperparameter tuning and improve reproducibility across models. Overall, NeuroML2Loihi demonstrates that standards-based neuromorphic neural simulation is feasible, efficient, and scientifically useful for small-scale biological models. More broadly, tools of this kind can reduce the software barrier between computational neuroscience and neuromorphic computing, enabling wider use of neuromorphic latency and energy advantages in neural simulation research. 

%% file: test.bib
@inproceedings{orchard2021efficient,
  title={Efficient neuromorphic signal processing with loihi 2},
  author={Orchard, Garrick and Frady, E Paxon and Rubin, Daniel Ben Dayan and Sanborn, Sophia and Shrestha, Sumit Bam and Sommer, Friedrich T and Davies, Mike},
  booktitle={2021 IEEE workshop on signal processing systems (SiPS)},
  pages={254--259},
  year={2021},
  organization={IEEE}
}

@article{van2001novel,
  title={A novel spike distance},
  author={van Rossum, Mark CW},
  journal={Neural computation},
  volume={13},
  number={4},
  pages={751--763},
  year={2001},
  publisher={MIT Press}
}

@article{victor1996nature,
  title={Nature and precision of temporal coding in visual cortex: a metric-space analysis},
  author={Victor, Jonathan D and Purpura, Keith P},
  journal={Journal of neurophysiology},
  volume={76},
  number={2},
  pages={1310--1326},
  year={1996},
  publisher={American Physiological Society Bethesda, MD}
}

@article{emmonsComprehensiveAnalysisElegans2024,
  title = {Comprehensive Analysis of the {{C}}. Elegans Connectome Reveals Novel Circuits and Functions of Previously Unstudied Neurons},
  author = {Emmons, Scott W.},
  year = 2024,
  month = dec,
  journal = {PLOS Biology},
  volume = {22},
  number = {12},
  pages = {e3002939},
  publisher = {Public Library of Science},
  issn = {1545-7885},
  doi = {10.1371/journal.pbio.3002939},
  urldate = {2026-02-23},
  langid = {english}
}

@article{heliasSupercomputersReadyUse2012,
  title = {Supercomputers {{Ready}} for {{Use}} as {{Discovery Machines}} for {{Neuroscience}}},
  author = {Helias, Moritz and Kunkel, Susanne and Masumoto, Gen and Igarashi, Jun and Eppler, Jochen Martin and Ishii, Shin and Fukai, Tomoki and Morrison, Abigail and Diesmann, Markus},
  year = 2012,
  month = nov,
  journal = {Frontiers in Neuroinformatics},
  volume = {6},
  pages = {26},
  issn = {1662-5196},
  doi = {10.3389/fninf.2012.00026},
  urldate = {2026-06-04},
  pmcid = {PMC3486988},
  pmid = {23129998}
}

@misc{kishoreSimplifiedKlinokinesisUsing2021,
  title = {Simplified {{Klinokinesis}} Using {{Spiking Neural Networks}} for {{Resource-Constrained Navigation}} on the {{Neuromorphic Processor Loihi}}},
  author = {Kishore, Apoorv and Saraswat, Vivek and Ganguly, Udayan},
  year = 2021,
  month = may,
  number = {arXiv:2105.01358},
  eprint = {2105.01358},
  primaryclass = {q-bio.NC},
  publisher = {arXiv},
  doi = {10.48550/arXiv.2105.01358},
  urldate = {2026-06-03},
  archiveprefix = {arXiv}
}

@article{kudithipudiNeuromorphicComputingScale2025a,
  title = {Neuromorphic Computing at Scale},
  author = {Kudithipudi, Dhireesha and Schuman, Catherine and Vineyard, Craig M. and Pandit, Tej and Merkel, Cory and Kubendran, Rajkumar and Aimone, James B. and Orchard, Garrick and Mayr, Christian and Benosman, Ryad and Hays, Joe and Young, Cliff and Bartolozzi, Chiara and Majumdar, Amitava and Cardwell, Suma George and Payvand, Melika and Buckley, Sonia and Kulkarni, Shruti and Gonzalez, Hector A. and Cauwenberghs, Gert and Thakur, Chetan Singh and Subramoney, Anand and Furber, Steve},
  year = 2025,
  month = jan,
  journal = {Nature},
  volume = {637},
  number = {8047},
  pages = {801--812},
  publisher = {Nature Publishing Group},
  issn = {1476-4687},
  doi = {10.1038/s41586-024-08253-8},
  urldate = {2026-03-11},
  copyright = {2025 Springer Nature Limited},
  langid = {english}
}

@misc{LargestNeuronalNetwork,
  title = {Largest Neuronal Network Simulation Achieved Using {{K}} Computer \textbar{} {{RIKEN}}},
  urldate = {2026-06-04},
  howpublished = {https://www.riken.jp/en/news\_pubs/research\_news/pr/2013/20130802\_1/}
}

@article{markramBlueBrainProject2006,
  title = {The {{Blue Brain Project}}},
  author = {Markram, Henry},
  year = 2006,
  month = feb,
  journal = {Nature Reviews Neuroscience},
  volume = {7},
  number = {2},
  pages = {153--160},
  publisher = {Nature Publishing Group},
  issn = {1471-0048},
  doi = {10.1038/nrn1848},
  urldate = {2026-04-05},
  copyright = {2006 Springer Nature Limited},
  langid = {english}
}

@misc{NextLevelNeuromorphicComputing,
  title = {Next-{{Level Neuromorphic Computing}}: {{Intel Lab}}'s {{Loihi}} 2 {{Chip}}},
  shorttitle = {Next-{{Level Neuromorphic Computing}}},
  journal = {Intel},
  urldate = {2026-06-09},
  howpublished = {https://www.intel.com/content/www/us/en/research/neuromorphic-computing-loihi-2-technology-brief.html},
  langid = {english}
}

@article{pedersenNeuromorphicIntermediateRepresentation2024,
  title = {Neuromorphic Intermediate Representation: {{A}} Unified Instruction Set for Interoperable Brain-Inspired Computing},
  shorttitle = {Neuromorphic Intermediate Representation},
  author = {Pedersen, Jens E. and Abreu, Steven and Jobst, Matthias and Lenz, Gregor and Fra, Vittorio and Bauer, Felix Christian and Muir, Dylan Richard and Zhou, Peng and Vogginger, Bernhard and Heckel, Kade and Urgese, Gianvito and Shankar, Sadasivan and Stewart, Terrence C. and Sheik, Sadique and Eshraghian, Jason K.},
  year = 2024,
  month = sep,
  journal = {Nature Communications},
  volume = {15},
  number = {1},
  pages = {8122},
  publisher = {Nature Publishing Group},
  issn = {2041-1723},
  doi = {10.1038/s41467-024-52259-9},
  urldate = {2025-12-07},
  copyright = {2024 The Author(s)},
  langid = {english}
}

@incollection{riddleMechanosensoryControlLocomotion1997a,
  title = {Mechanosensory {{Control}} of {{Locomotion}}},
  booktitle = {C. Elegans {{II}}. 2nd Edition},
  author = {Riddle, Donald L. and Blumenthal, Thomas and Meyer, Barbara J. and Priess, James R.},
  year = 1997,
  publisher = {Cold Spring Harbor Laboratory Press},
  urldate = {2026-04-17},
  langid = {english}
}

@article{brennerNaturesGiftScience2003,
  title = {Nature's {{Gift}} to {{Science}} ({{Nobel Lecture}})},
  author = {Brenner, Sydney},
  year = 2003,
  month = aug,
  journal = {ChemBioChem},
  volume = {4},
  number = {8},
  pages = {683--687},
  issn = {1439-4227, 1439-7633},
  doi = {10.1002/cbic.200300625},
  urldate = {2025-12-13},
  langid = {english}
}

@article{varshneyStructuralPropertiesCaenorhabditis2011,
  title = {Structural {{Properties}} of the {{Caenorhabditis}} Elegans {{Neuronal Network}}},
  author = {Varshney, Lav R. and Chen, Beth L. and Paniagua, Eric and Hall, David H. and Chklovskii, Dmitri B.},
  editor = {Sporns, Olaf},
  year = 2011,
  month = feb,
  journal = {PLoS Computational Biology},
  volume = {7},
  number = {2},
  pages = {e1001066},
  issn = {1553-7358},
  doi = {10.1371/journal.pcbi.1001066},
  urldate = {2026-06-11},
  langid = {english}
}

@article{yamazakiHumanscaleBrainSimulation2021,
  title = {Human-Scale {{Brain Simulation}} via {{Supercomputer}}: {{A Case Study}} on the {{Cerebellum}}},
  shorttitle = {Human-Scale {{Brain Simulation}} via {{Supercomputer}}},
  author = {Yamazaki, Tadashi and Igarashi, Jun and Yamaura, Hiroshi},
  year = 2021,
  month = may,
  journal = {Neuroscience},
  volume = {462},
  pages = {235--246},
  publisher = {Elsevier},
  issn = {0306-4522, 1873-7544},
  doi = {10.1016/j.neuroscience.2021.01.014},
  urldate = {2026-06-04},
  langid = {english}
}

@article{chalfieNeuralCircuitTouch1985,
  title = {The Neural Circuit for Touch Sensitivity in {{Caenorhabditis}} Elegans},
  author = {Chalfie, M and Sulston, {\relax JE} and White, {\relax JG} and Southgate, E and Thomson, {\relax JN} and Brenner, S},
  year = 1985,
  month = apr,
  journal = {The Journal of Neuroscience},
  volume = {5},
  number = {4},
  pages = {956--964},
  issn = {0270-6474},
  doi = {10.1523/JNEUROSCI.05-04-00956.1985},
  urldate = {2026-06-12},
  pmcid = {PMC6565008},
  pmid = {3981252}
}

@conference{elephant18,
    author = {Denker, M. and Yegenoglu, A. and Grün, S.},
    booktitle = {Neuroinformatics 2018},
    title = {{C}ollaborative {HPC}-enabled workflows on the {HBP} {C}ollaboratory using the {E}lephant framework},
    pages = {P19},
    year = {2018},
    doi = {10.12751/incf.ni2018.0019},
    url = {https://abstracts.g-node.org/conference/NI2018/abstracts#/uuid/023bec4e-0c35-4563-81ce-2c6fac282abd},
}

@inproceedings{hagbergExploringNetworkStructure2008,
  title = {Exploring {{Network Structure}}, {{Dynamics}}, and {{Function}} Using {{NetworkX}}},
  booktitle = {Python in {{Science Conference}}},
  author = {Hagberg, Aric A. and Schult, Daniel A. and Swart, Pieter J.},
  year = 2008,
  month = jun,
  pages = {11--15},
  address = {Pasadena, California},
  doi = {10.25080/TCWV9851},
  urldate = {2026-06-12},
  copyright = {https://creativecommons.org/licenses/by/3.0/}
}

@article{szigeti2014openworm,
  author  = {Szigeti, Bal{\'a}zs and Gleeson, Padraig and Vella, Michael and
             Khayrulin, Sergey and Palyanov, Andrey and Hokanson, Jim and
             Currie, Michael and Cantarelli, Matteo and Idili, Giovanni and
             Larson, Stephen},
  title   = {{OpenWorm}: An Open-Science Approach to Modeling
             \emph{Caenorhabditis elegans}},
  journal = {Frontiers in Computational Neuroscience},
  volume  = {8},
  pages   = {137},
  year    = {2014},
  doi     = {10.3389/fncom.2014.00137},
}

@article{gleeson2018c302,
  author  = {Gleeson, Padraig and Lung, David and Grosu, Radu and
             Hasani, Ramin and Larson, Stephen D.},
  title   = {{c302}: A Multiscale Framework for Modelling the Nervous System
             of \emph{Caenorhabditis elegans}},
  journal = {Philosophical Transactions of the Royal Society B},
  volume  = {373},
  number  = {1758},
  pages   = {20170379},
  year    = {2018},
  doi     = {10.1098/rstb.2017.0379},
}

@article{gleeson2010neuroml,
  author  = {Gleeson, Padraig and Crook, Sharon and Cannon, Robert C. and
             Hines, Michael L. and Billings, Guy O. and Farinella, Matteo and
             Morse, Thomas M. and Davison, Andrew P. and Ray, Subhasis and
             Bhalla, Upinder S. and Barnes, Simon R. and Dimitrova, Yoana D. and
             Silver, R. Angus},
  title   = {{NeuroML}: A Language for Describing Data Driven Models of Neurons
             and Networks with a High Degree of Biological Detail},
  journal = {PLoS Computational Biology},
  volume  = {6},
  number  = {6},
  pages   = {e1000815},
  year    = {2010},
  doi     = {10.1371/journal.pcbi.1000815},
}

@article{hinesNEURONSimulationEnvironment1997,
  title = {The {{NEURON Simulation Environment}}},
  author = {Hines, M. L. and Carnevale, N. T.},
  year = 1997,
  month = aug,
  journal = {Neural Computation},
  volume = {9},
  number = {6},
  pages = {1179--1209},
  issn = {0899-7667, 1530-888X},
  doi = {10.1162/neco.1997.9.6.1179},
  urldate = {2026-06-27},
  langid = {english}
}

@article{kumbharCoreNEURONOptimizedCompute2019,
  title = {{{CoreNEURON}} : {{An Optimized Compute Engine}} for the {{NEURON Simulator}}},
  shorttitle = {{{CoreNEURON}}},
  author = {Kumbhar, Pramod and Hines, Michael and Fouriaux, Jeremy and Ovcharenko, Aleksandr and King, James and Delalondre, Fabien and Sch{\"u}rmann, Felix},
  year = 2019,
  month = sep,
  journal = {Frontiers in Neuroinformatics},
  volume = {13},
  publisher = {Frontiers},
  issn = {1662-5196},
  doi = {10.3389/fninf.2019.00063},
  urldate = {2026-06-15},
  langid = {english}
}

@article{vellaLibNeuroMLPyLEMSUsing2014,
  title = {{{libNeuroML}} and {{PyLEMS}}: Using {{Python}} to Combine Procedural and Declarative Modeling Approaches in Computational Neuroscience},
  shorttitle = {{{libNeuroML}} and {{PyLEMS}}},
  author = {Vella, Michael and Cannon, Robert C. and Crook, Sharon and Davison, Andrew P. and Ganapathy, Gautham and Robinson, Hugh P. C. and Silver, R. Angus and Gleeson, Padraig},
  year = 2014,
  month = apr,
  journal = {Frontiers in Neuroinformatics},
  volume = {8},
  issn = {1662-5196},
  doi = {10.3389/fninf.2014.00038},
  urldate = {2026-06-13}
}

@inproceedings{gaurav2024legendre,
  title={Legendre-snn on Loihi-2: Evaluation and insights},
  author={Gaurav, Ramashish and Stewart, Terrence C and Yi, Yang},
  booktitle={NeurIPS 2024 Workshop Machine Learning with new Compute Paradigms},
  year={2024}
}

@article{kunkelSpikingNetworkSimulation2014,
  title = {Spiking Network Simulation Code for Petascale Computers},
  author = {Kunkel, Susanne and Schmidt, Maximilian and Eppler, Jochen M. and Plesser, Hans E. and Masumoto, Gen and Igarashi, Jun and Ishii, Shin and Fukai, Tomoki and Morrison, Abigail and Diesmann, Markus and Helias, Moritz},
  year = 2014,
  month = oct,
  journal = {Frontiers in Neuroinformatics},
  volume = {8},
  pages = {78},
  issn = {1662-5196},
  doi = {10.3389/fninf.2014.00078},
  urldate = {2026-06-27},
  pmcid = {PMC4193238},
  pmid = {25346682}
}

@article{gleesonNeuroMLLanguageDescribing2010a,
  title = {{{NeuroML}}: {{A Language}} for {{Describing Data Driven Models}} of {{Neurons}} and {{Networks}} with a {{High Degree}} of {{Biological Detail}}},
  shorttitle = {{{NeuroML}}},
  author = {Gleeson, Padraig and Crook, Sharon and Cannon, Robert C. and Hines, Michael L. and Billings, Guy O. and Farinella, Matteo and Morse, Thomas M. and Davison, Andrew P. and Ray, Subhasis and Bhalla, Upinder S. and Barnes, Simon R. and Dimitrova, Yoana D. and Silver, R. Angus},
  editor = {Friston, Karl J.},
  date = {2010-06-17},
  journaltitle = {PLoS Computational Biology},
  shortjournal = {PLoS Comput Biol},
  volume = {6},
  number = {6},
  pages = {e1000815},
  issn = {1553-7358},
  doi = {10.1371/journal.pcbi.1000815},
  url = {https://dx.plos.org/10.1371/journal.pcbi.1000815},
  urldate = {2025-12-27},
  langid = {english}
}

@article{whiteStructureNervousSystem1986,
  title = {The Structure of the Nervous System of the Nematode {{{\emph{Caenorhabditis}}}}{\emph{ Elegans}}},
  author = {White, John Graham and Southgate, Eileen and Thomson, J. N. and Brenner, Sydney},
  year = 1986,
  month = nov,
  journal = {Philosophical Transactions of the Royal Society of London. B, Biological Sciences},
  volume = {314},
  number = {1165},
  pages = {1--340},
  issn = {0080-4622, 2054-0280},
  doi = {10.1098/rstb.1986.0056},
  urldate = {2026-07-05},
  copyright = {https://royalsociety.org/journals/ethics-policies/data-sharing-mining/},
  langid = {english}
}

@article{indiveriImportanceSpaceTime2019,
  title = {The Importance of Space and Time in Neuromorphic Cognitive Agents},
  author = {Indiveri, Giacomo and Sandamirskaya, Yulia},
  year = 2019,
  month = nov,
  journal = {IEEE Signal Processing Magazine},
  volume = {36},
  number = {6},
  eprint = {1902.09791},
  primaryclass = {cs},
  pages = {16--28},
  issn = {1053-5888, 1558-0792},
  doi = {10.1109/MSP.2019.2928376},
  urldate = {2026-03-16},
  archiveprefix = {arXiv}
}

@article{schreiberEmulatingInsectBrains2024,
  title = {Emulating Insect Brains for Neuromorphic Navigation},
  author = {Schreiber, Korbinian and Wunderlich, Timo and Spilger, Philipp and Billaudelle, Sebastian and Cramer, Benjamin and Stradmann, Yannik and Pehle, Christian and M{\"u}ller, Eric and Petrovici, Mihai A. and Schemmel, Johannes and Meier, Karlheinz},
  year = 2024,
  publisher = {arXiv},
  doi = {10.48550/ARXIV.2401.00473},
  urldate = {2026-06-27},
  copyright = {arXiv.org perpetual, non-exclusive license}
}

@article{vanalbadaPerformanceComparisonDigital2018,
  title = {Performance {{Comparison}} of the {{Digital Neuromorphic Hardware SpiNNaker}} and the {{Neural Network Simulation Software NEST}} for a {{Full-Scale Cortical Microcircuit Model}}},
  author = {{van Albada}, Sacha J. and Rowley, Andrew G. and Senk, Johanna and Hopkins, Michael and Schmidt, Maximilian and Stokes, Alan B. and Lester, David R. and Diesmann, Markus and Furber, Steve B.},
  year = 2018,
  month = may,
  journal = {Frontiers in Neuroscience},
  volume = {12},
  publisher = {Frontiers},
  issn = {1662-453X},
  doi = {10.3389/fnins.2018.00291},
  urldate = {2026-06-27},
  langid = {english}
}

@article{rhodesRealtimeCorticalSimulation2019,
  title = {Real-Time Cortical Simulation on Neuromorphic Hardware},
  author = {Rhodes, Oliver and Peres, Luca and Rowley, Andrew G. D. and Gait, Andrew and Plana, Luis A. and Brenninkmeijer, Christian and Furber, Steve B.},
  year = 2019,
  month = dec,
  journal = {Philosophical Transactions of the Royal Society A: Mathematical, Physical and Engineering Sciences},
  volume = {378},
  number = {2164},
  pages = {20190160},
  issn = {1364-503X},
  doi = {10.1098/rsta.2019.0160},
  urldate = {2026-06-27}
}

@article{dorkenwaldNeuronalWiringDiagram2024,
  title = {Neuronal Wiring Diagram of an Adult Brain},
  author = {Dorkenwald, Sven and Matsliah, Arie and Sterling, Amy R. and Schlegel, Philipp and Yu, Szi-chieh and McKellar, Claire E. and Lin, Albert and Costa, Marta and Eichler, Katharina and Yin, Yijie and Silversmith, Will and {Schneider-Mizell}, Casey and Jordan, Chris S. and Brittain, Derrick and Halageri, Akhilesh and Kuehner, Kai and Ogedengbe, Oluwaseun and Morey, Ryan and Gager, Jay and Kruk, Krzysztof and Perlman, Eric and Yang, Runzhe and Deutsch, David and Bland, Doug and Sorek, Marissa and Lu, Ran and Macrina, Thomas and Lee, Kisuk and Bae, J. Alexander and Mu, Shang and Nehoran, Barak and Mitchell, Eric and Popovych, Sergiy and Wu, Jingpeng and Jia, Zhen and Castro, Manuel A. and Kemnitz, Nico and Ih, Dodam and Bates, Alexander Shakeel and Eckstein, Nils and Funke, Jan and Collman, Forrest and Bock, Davi D. and Jefferis, Gregory S. X. E. and Seung, H. Sebastian and Murthy, Mala},
  year = 2024,
  month = oct,
  journal = {Nature},
  volume = {634},
  number = {8032},
  pages = {124--138},
  publisher = {Nature Publishing Group},
  issn = {1476-4687},
  doi = {10.1038/s41586-024-07558-y},
  urldate = {2026-06-28},
  copyright = {2024 The Author(s)},
  langid = {english}
}

@misc{wangNeuromorphicSimulationDrosophila2025,
  title = {Neuromorphic {{Simulation}} of {{Drosophila Melanogaster Brain Connectome}} on {{Loihi}} 2},
  author = {Wang, Felix and Theilman, Bradley H. and Rothganger, Fred and Severa, William and Vineyard, Craig M. and Aimone, James B.},
  year = 2025,
  publisher = {arXiv},
  doi = {10.48550/ARXIV.2508.16792},
  urldate = {2025-12-13},
  copyright = {Creative Commons Attribution 4.0 International}
}

@article{gewaltigNESTNEuralSimulation2007a,
  title = {{{NEST}} ({{NEural Simulation Tool}})},
  author = {Gewaltig, Marc-Oliver and Diesmann, Markus},
  year = 2007,
  journal = {Scholarpedia},
  volume = {2},
  number = {4},
  pages = {1430},
  issn = {1941-6016},
  doi = {10.4249/scholarpedia.1430},
  urldate = {2026-07-05},
  langid = {english}
}

@article{goodmanBrianSimulatorSpiking2008,
  title = {Brian: A Simulator for Spiking Neural Networks in {{Python}}},
  shorttitle = {Brian},
  author = {Goodman, Dan},
  year = 2008,
  journal = {Frontiers in Neuroinformatics},
  volume = {2},
  issn = {16625196},
  doi = {10.3389/neuro.11.005.2008},
  urldate = {2026-07-05}
}

@article{stimbergBrian2Intuitive2019a,
  title = {Brian 2, an Intuitive and Efficient Neural Simulator},
  author = {Stimberg, Marcel and Brette, Romain and Goodman, Dan Fm},
  year = 2019,
  month = aug,
  journal = {eLife},
  volume = {8},
  pages = {e47314},
  issn = {2050-084X},
  doi = {10.7554/eLife.47314},
  urldate = {2026-07-05},
  copyright = {http://creativecommons.org/licenses/by/4.0/},
  langid = {english}
}

@misc{CUDAProgrammingGuide,
  title = {{{CUDA Programming Guide}} --- {{CUDA Programming Guide}}},
  urldate = {2026-06-28},
  howpublished = {https://docs.nvidia.com/cuda/cuda-programming-guide/index.html}
}

@misc{AMDROCmDocumentation,
  title = {{{AMD ROCm}} Documentation --- {{ROCm Documentation}}},
  urldate = {2026-06-28},
  howpublished = {https://rocm.docs.amd.com/en/latest/}
}

@article{davisonPyNNCommonInterface2008,
  title = {{{PyNN}}: A Common Interface for Neuronal Network Simulators},
  shorttitle = {{{PyNN}}},
  author = {Davison, Andrew P},
  year = 2008,
  journal = {Frontiers in Neuroinformatics},
  volume = {2},
  issn = {16625196},
  doi = {10.3389/neuro.11.011.2008},
  urldate = {2026-06-27}
}

@article{rothgangerN2AComputationalTool2014,
  title = {{{N2A}}: A Computational Tool for Modeling from Neurons to Algorithms},
  shorttitle = {{{N2A}}},
  author = {Rothganger, Fredrick and Warrender, Christina E. and Trumbo, Derek and Aimone, James B.},
  year = 2014,
  journal = {Frontiers in Neural Circuits},
  volume = {8},
  issn = {1662-5110},
  doi = {10.3389/fncir.2014.00001},
  urldate = {2026-06-26}
}

@misc{NengoLoihiNengoLoihi111dev0,
  title = {{{NengoLoihi}} --- {{NengoLoihi}} 1.1.1.Dev0 Docs},
  urldate = {2026-06-28},
  howpublished = {https://www.nengo.ai/nengo-loihi/}
}

@misc{generSNNMLIRMLIRDialect2026,
  title = {{{SNN-MLIR}}: {{An MLIR Dialect}} for {{Compiling Neuromorphic SNNs}} from {{NIR}} to {{Bare-Metal C}}},
  shorttitle = {{{SNN-MLIR}}},
  author = {Gener, Alejandro Garc{\'i}a and de Pinedo, Alvaro Roll{\'o}n},
  year = 2026,
  month = jun,
  number = {arXiv:2606.09213},
  eprint = {2606.09213},
  primaryclass = {cs.PL},
  publisher = {arXiv},
  doi = {10.48550/arXiv.2606.09213},
  urldate = {2026-06-27},
  archiveprefix = {arXiv}
}

@misc{negriBrian2Lava2025,
  title = {{{Brian2Lava}}},
  author = {Negri, Francesco and Luboeinski, Jannik* and Michaelis, Carlo and Oed, Winfried and St{\"o}ber, Tristan M. and Lehr, Andrew B. and Tetzlaff, Christian},
  year = 2025,
  month = mar,
  doi = {10.5281/ZENODO.18470808},
  urldate = {2026-07-15},
  copyright = {MIT License},
  howpublished = {Zenodo}
}

@article{furberOverviewSpiNNakerSystem2013,
  title = {Overview of the {{SpiNNaker System Architecture}}},
  author = {Furber, Steve B. and Lester, David R. and Plana, Luis A. and Garside, Jim D. and Painkras, Eustace and Temple, Steve and Brown, Andrew D.},
  year = 2013,
  month = dec,
  journal = {IEEE Transactions on Computers},
  volume = {62},
  number = {12},
  pages = {2454--2467},
  issn = {0018-9340},
  doi = {10.1109/TC.2012.142},
  urldate = {2026-07-15},
  copyright = {https://ieeexplore.ieee.org/Xplorehelp/downloads/license-information/IEEE.html}
}

@misc{lava,
  title        = {Lava: A Software Framework for Neuromorphic Computing},
  author       = {{Intel Neuromorphic Computing Lab}},
  year         = {2024},
  howpublished = {\url{https://lava-nc.org}},
  note         = {Accessed: 2026-07-16}
}

@online{dannyrosenDRosen766NeuroML2LoihiNeuroML2Loihi2026,
  author = {Danny Rosen},
  title = {NeuroML2Loihi},
  year = {2026},
  month = jul,
  version = {1.0.0},
  publisher = {Zenodo},
  doi = {10.5281/zenodo.21431945},
  url = {https://github.com/DRosen766/NeuroML2Loihi},
}

@article{michaelisBrian2LoihiEmulatorNeuromorphic2022,
  title = {{{Brian2Loihi}}: {{An}} Emulator for the Neuromorphic Chip {{Loihi}} Using the Spiking Neural Network Simulator {{Brian}}},
  shorttitle = {{{Brian2Loihi}}},
  author = {Michaelis, Carlo and Lehr, Andrew B. and Oed, Winfried and Tetzlaff, Christian},
  year = 2022,
  month = nov,
  journal = {Frontiers in Neuroinformatics},
  volume = {16},
  pages = {1015624},
  issn = {1662-5196},
  doi = {10.3389/fninf.2022.1015624},
  urldate = {2026-07-18}
}

@article{hobertRevisitingNeuronalCell2016,
  title = {Revisiting {{Neuronal Cell Type Classification}} in {{Caenorhabditis}} Elegans},
  author = {Hobert, Oliver and Glenwinkel, Lori and White, John},
  year = 2016,
  month = nov,
  journal = {Current Biology},
  volume = {26},
  number = {22},
  pages = {R1197-R1203},
  issn = {09609822},
  doi = {10.1016/j.cub.2016.10.027},
  urldate = {2026-07-18},
  langid = {english}
}
